Original Paper

# A Patient Simulation Framework for Risk Assessment of Conversational Healthcare AI: Evaluation of an Antidepressant Decision Aid

Md Tanvir Rouf Shawon[1]; Mohammad Sabik Irbaz[2]; Hadeel R. A. Elyazori[2]; Keerti Reddy Resapu[3]; Yili Lin[3]; Vladimir Franzuela Cardenas[3]; K. Pierre Eklou[4]; Farrokh Alemi[3]; Kevin Lybarger[2]

[1]Department of Computer Science, George Mason University, Fairfax, VA, United States

[2]Department of Information Sciences and Technology, George Mason University, Fairfax, VA, United States

[3]Department of Health Administration and Policy, George Mason University, Fairfax, VA, United States

[4]School of Nursing, College of Public Health, George Mason University, Fairfax, VA, United States

**Corresponding Author:**
Md Tanvir Rouf Shawon
College of Engineering and Computing
George Mason University
4400 University Dr. MSN 1G8
Fairfax, VA, 22030
United States
Phone: 1 5716635183
Email: mshawon@gmu.edu

## Abstract

**Background:** Conversational AI systems are increasingly deployed in healthcare for clinical decision support, but their performance varies substantially across patient communication styles, health literacy levels, and behavioral patterns. Static benchmarks cannot capture multi-turn dynamics through which this variation compounds, and no current evaluation framework implements structured AI risk management guidance for conversational healthcare AI. The result is a structural risk: AI systems may perform well in aggregate while failing disproportionately for the populations they are intended to help.

**Objective:** This study develops and validates a patient simulation framework that aligns with the National Institute of Standards and Technology (NIST) AI Risk Management Framework (AI RMF) MAP and MEASURE functions, providing an empirical basis for identifying and characterizing performance risks in conversational clinical AI across medical, linguistic, and behavioral patient variation.

We applied the framework to a conversational decision aid for antidepressant selection in major depressive disorder (the AI Decision Aid).

**Methods:** The simulator integrates three profile dimensions: (1) medical profiles constructed from All of Us electronic health records using risk-ratio gating; (2) linguistic profiles modeling a health literacy gradient and condition-specific communication; and (3) behavioral profiles representing cooperative, distracted, and adversarial engagement. We generated 500 simulated conversations and evaluated profile fidelity through human annotation and a large language model (LLM) judge, then assessed downstream effects on the AI Decision Aid's concept retrieval and antidepressant recommendations.

**Results:** The patient simulator expressed medical concepts with high fidelity (96.6% accurate across 8,210 concepts), with human inter-annotator agreement of 0.73 κ and LLM-judge agreement against human annotators of 0.78 κ, both indicating substantial agreement. Behavioral profiles were reliably distinguished (0.93 κ, near perfect agreement), and linguistic profiles showed substantial agreement at the lower bound of the substantial range (0.61 κ), adequate to support profile-level analysis. The framework revealed monotonic degradation in AI Decision Aid performance across the health literacy gradient. Rank-1 concept retrieval increased from 47.6% for limited health literacy to 81.9% for proficient health literacy, with corresponding declines in antidepressant recommendation accuracy.

**Conclusions:** Patient simulation grounded in the NIST AI RMF exposes measurable performance risks in conversational healthcare AI that static benchmarks miss, with direct equity implications. Health literacy operates as a structural risk factor, with degraded performance concentrated in patients carrying the greatest burden of psychiatric illness. The framework supports targeted risk-mitigation interventions before deployment. Though we evaluated it only on antidepressant selection, extending the open-source framework to other clinical decision-aid tasks is future work.

## Introduction

Conversational AI systems are increasingly deployed in healthcare to support clinical decision-making, patient engagement, and care accessibility [1–4]. Yet these systems face a fundamental evaluation gap: patients communicate the same clinical information in vastly different ways depending on health literacy, psychological state, and interaction style [5,6]. A patient who says, "my morning pill for my nerves" and one who reports "20 mg of fluoxetine for generalized anxiety disorder" may describe identical clinical realities, but a conversational agent that handles the second better than the first introduces a structural bias against vulnerable populations. Recent work demonstrates that large language model (LLM) clinical outputs shift measurably when patient messages are perturbed with stylistic, syntactic, or demographic variation, with disparities concentrated in vulnerable subgroups and amplified in conversational settings [7]. There is no standardized method to assess whether AI performance remains equitable across this variation [8,9], so disparities are likely to surface only after deployment, with disproportionate impact on populations facing existing barriers to equitable care. LLM-based approaches lower barriers to building such systems [10,11] but heighten the need for rigorous, scalable risk assessment that can characterize performance across diverse patient communication patterns.

Static benchmarks cannot capture these risks because they bypass the multi-turn dynamics through which communication variation compounds [12,13]. Patient simulation offers a scalable alternative: simulated patients with controlled clinical, linguistic, and behavioral characteristics can systematically probe system performance across populations that would be difficult, slow, or ethically complex to recruit for live evaluation [14]. Simulation-based methods are also being used more broadly to quantify LLM risk in healthcare, for example to estimate hazard-to-harm probabilities for regulatory risk assessment of LLM-based medical devices [15]. What is missing is a structured framework for connecting patient simulation to AI risk governance. The National Institute of Standards and Technology (NIST) AI Risk Management Framework (AI RMF) provides domain-agnostic guidance for identifying, assessing, and managing AI risks [16], but no existing patient simulation framework operationalizes this guidance for conversational healthcare AI [8,9]. This work introduces such a framework. Patient simulator construction supports the MAP function by structuring risk identification across medical, linguistic, and behavioral profile dimensions. Simulator-AI interaction supports the MEASURE function by surfacing performance variation across the structured profile space. The framework surfaces the evidence needed for AI RMF-aligned risk analysis. Extending this evidence into deployment-specific risk analyses is left to future work.

We present a patient simulator grounded in real-world clinical data from the All of Us Research Program [17], evaluated on a conversational decision aid for antidepressant selection in major depressive disorder (MDD). The simulator integrates three profile dimensions: (1) medical profiles constructed from electronic health record (EHR) data through a risk-ratio-based feature selection process that prioritizes outcome-relevant clinical features while preserving statistical independence and clinical coherence; (2) linguistic profiles modeling health literacy variation [18,19] and condition-specific communication patterns [20,21]; and (3) behavioral profiles representing empirically derived interaction patterns including cooperative, distracted, and adversarial engagement [22,23]. The framework generalizes across healthcare tasks and conditions; this study evaluates it on antidepressant selection while also addressing the following research questions:

- **RQ1.** How effectively can structured electronic health record (EHR) data be transformed into risk-aware medical profiles aligned with AI trustworthiness principles?

- **RQ2.** How can simulated patients combining medical, linguistic, and behavioral information produce realistic and distinguishable conversational behavior?
- **RQ3.** How does simulated patient variation across medical, linguistic, and behavioral profiles affect the performance of a conversational clinical decision aid, and which patient subgroups are most affected?

This work contributes: (1) a patient simulation framework that aligns with the NIST AI RMF MAP and MEASURE functions for conversational healthcare AI, integrating medical, linguistic, and behavioral profiles, (2) empirical evidence that health literacy creates a monotonic performance gradient in conversational clinical AI, identifying a concrete equity risk, (3) a risk-ratio-based algorithm for generating outcome-relevant, auditable medical profiles from EHR data, and (4) a controlled perturbation methodology using ontology-aware semantic substitution for validating both human annotators and LLM judges. The work was conducted by a multidisciplinary team integrating clinical psychiatric and mental health nursing expertise, clinical informatics, and natural language processing, with guidance from a patient and clinician advisory board. The code, source data, and simulated interactions are publicly available [24].

## Related Work

In this section, we review related work on user simulation, patient simulation, and behavioral and linguistic modeling for conversational AI.

User simulation provides scalable methods for evaluating conversational agents across domains; patient simulation adapts these methods for clinical dialogue and diagnostic reasoning.

### *User Simulation*

User simulation enables controlled, scalable evaluation of interactive systems without requiring human trials. Foundational surveys span information access, dialogue modeling, and recommendation systems [14,25–27], while recent work leverages LLMs to generate context-aware user behavior through dual-model architectures combining generators and verifiers [28,29] and session-level search simulation [30,31]. In healthcare, synthetic users with clinical profiles evaluate decision support and coaching systems [32], establishing user simulation as a critical method for assessing reliability and safety in AI systems. Mental health conversational AI research has historically split between technical evaluation of response quality and clinical evaluation of patient outcomes, with limited integration across the two [33].

### *Patient Simulation*

Patient simulation involves construction of virtual patients that engage in interactive clinical dialogues with human or automated clinicians [34]. Frameworks vary widely in patient state representation, from hand-crafted profiles to EHR-grounded models, and in how they control disclosure, tone, and clinical accuracy [1,35,36]. Prior work often treats patient simulation as a single unified problem, but it involves two distinct technical challenges: (1) constructing clinically valid patient representations and (2) simulating interactive dialogue that expresses those representations over time [36]. Recent systems have integrated LLMs to enhance expressiveness and flexibility [37], constructed structured knowledge bases from clinical notes to support medical intake tasks [38], and embedded risk-aware feedback mechanisms [34].

### *Data-Driven Medical Realism*

Early systems employed deterministic state machines or probabilistic sequence models to generate medical histories. Synthea [39] and SynSys [40] produce scalable, transparent timelines but reflect population-level distributions rather than individual patient data. EHR-grounded approaches improve realism. Generative Adversarial Network (GAN) variants [41,42] address missing features and mixed data types, achieving higher fidelity than rule-based methods on MIMIC-III/IV benchmarks [43], while SimSUM [44] combines Bayesian networks with prompted LLMs to synthesize clinical notes. Systems integrating EHR grounding with conversational retrieval, including ophthalmology simulators [38] and multi-agent knowledge graph pipelines [45], advance clinical accuracy but treat the encounter as a transactional data exchange rather than a dynamic, rapport-dependent interaction.

### *Behavioral and Linguistic Modeling*

Beyond medical profile construction, systems must express profiles through natural dialogue with appropriate communication style and conversational dynamics [35,46]. Cognitive and persona-based approaches model psychological states to generate behaviorally coherent conversation. PATIENT Ψ [47] uses expert-designed cognitive schemas with GPT-4 to reproduce emotional fluctuations and resistant behaviors, while SFMSS [48] embeds Big Five traits to shape dialogue tone and coordinates patient, nurse, and supervisor agents to enforce outpatient triage workflows. PAL [49] simulates emotionally nuanced palliative care patient interactions with NURSE-framework feedback, illustrating affect-grounded simulation in another condition-specific clinical context. These systems remain condition-specific (e.g., mental health, palliative care) and lack medical grounding for clinical evaluation. Prompt-based approaches [34,37] offer scalability but lack persistent patient state tracking and systematic control over linguistic and behavioral variation.

### *Procedural and Workflow Control*

A third class prioritizes procedural fidelity and clinical workflows through discrete agent roles and actions. iPDG [50] enforces clinical plausibility through manually defined domain rules. Such systems achieve procedural coherence but sacrifice conversational flexibility and behavioral depth.

### *Limitations and Gaps*

Recent systematic reviews of mental health chatbots find that LLM-based systems concentrate in early-stage technical validation, with most evaluations focused on conversational quality or adherence to specific prompts in controlled settings rather than rigorous testing for clinical benefit [51]. Existing patient simulation systems exhibit two critical gaps that constrain trustworthy AI evaluation. First, no existing framework integrates medical realism, behavioral variation, and linguistic diversity within a unified architecture. Recent systems pair medical grounding with behavioral modeling or behavioral depth with linguistic variation, but this fragmentation creates evaluation blind spots. Existing systems cannot systematically test whether agents maintain diagnostic accuracy across complex comorbidities, varied health literacy, and adversarial behavior simultaneously. Multi-turn failures like context loss and inconsistent safety emerge precisely at these intersections.

Second, existing approaches prioritize medical outcomes over systematic risk management aligned with frameworks like the NIST AI RMF. They lack explicit risk mapping connecting simulation parameters to risk categories, auditability with traceable lineage to source data, and controllable risk probing through systematic variation. This prevents detection of consequential failures: agents may recommend different treatments when patients express identical conditions at different health literacy levels, or safety mechanisms may fail under adversarial conditions that only systematic variation can expose. We address both gaps through a unified framework whose medical, linguistic, and behavioral profiles align with the NIST AI RMF MAP and MEASURE functions, enabling risk assessment across medical accuracy, communication appropriateness, and behavioral robustness.

## Methods

### Overview

The framework integrates a patient simulator with an *AI Decision Aid* to enable systematic evaluation of conversational clinical decision-making. The patient simulator produces controlled, profile-driven responses by combining medical, linguistic, and behavioral characteristics, while the *AI Decision Aid* conducts a structured conversational intake to elicit clinical history and generate antidepressant recommendations. Figure 1 illustrates this interaction. This section details the patient simulator design, *AI Decision Aid*, and evaluation methodology.

The research team integrates clinical expertise in psychiatric and mental health nursing with expertise in clinical informatics, health administration, and natural language processing, with clinical authors contributing to the patient profiles, antidepressant selection task, and evaluation framework. The broader Patient-Centered Outcomes Research Institute (PCORI)-funded project is guided by an advisory board of clinicians, mental health organization leaders, and individuals with lived experience of depression.

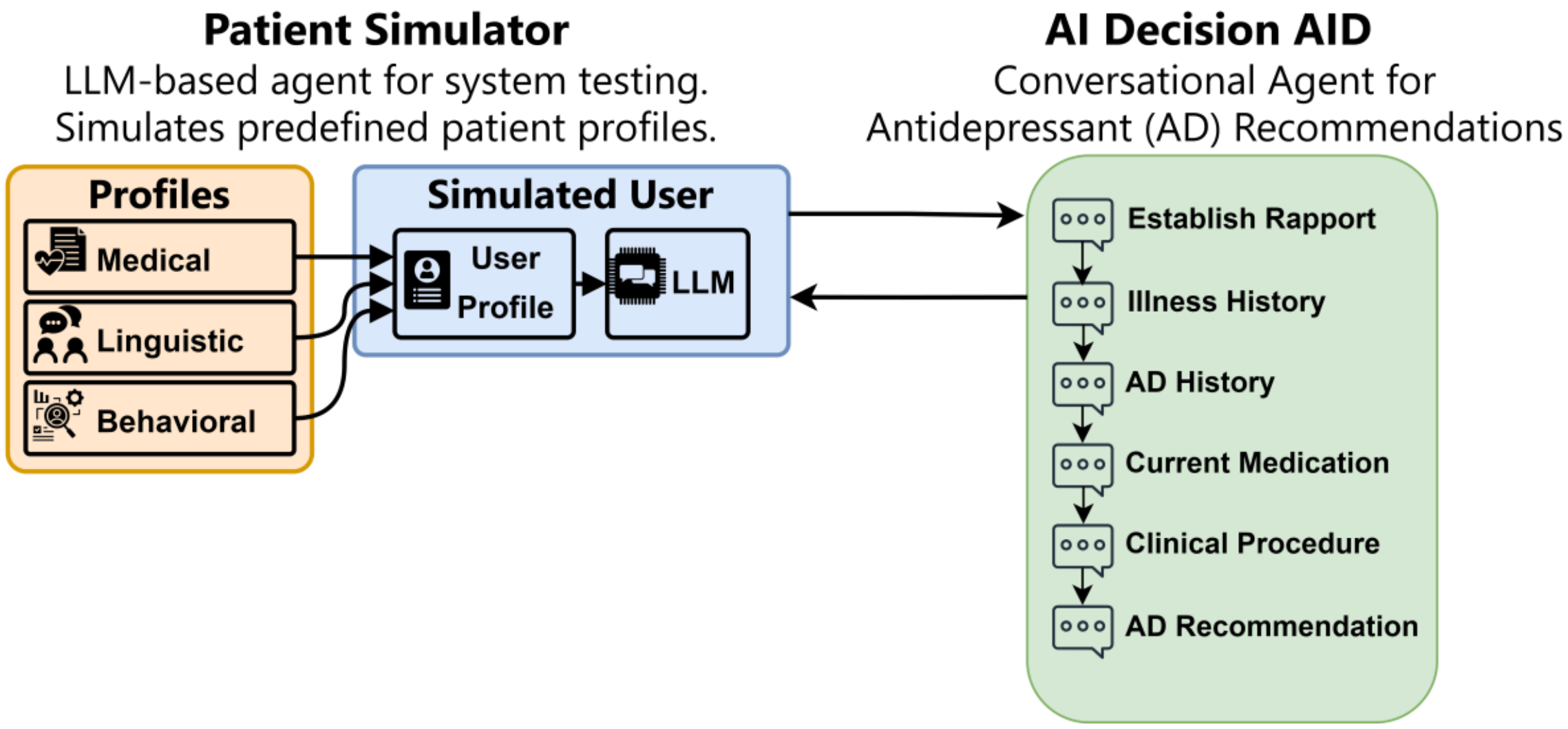

Figure 1: A schematic diagram of the conversation between patient simulator and the *AI Decision Aid* system.

## Data

Both the patient simulator medical profiles and the *AI Decision Aid* prediction models are derived from the All of Us Research Program Registered Tier v8 dataset, a national longitudinal study providing structured EHR data including conditions, medications, procedures, and demographics [17].

### *Cohort Selection*

We extracted data from All of Us participants with major depressive disorder (Systematized Nomenclature of Medicine Clinical Terms (SNOMED CT) Code 370143000 and its descendants). The resulting cohort includes 58,446 participants, contributing 466,752 antidepressant trials, with 18,471 diagnoses, 2,642 medications, and 5,001 procedures. Outcomes capture responses to 14 antidepressants and one category covering all remaining antidepressants (dataset cut-off: October 1, 2023). Antidepressant response was defined as taking the antidepressant for at least 10 weeks without switching or augmenting with another antidepressant [52].

### *Data Application*

For patient simulator medical profiles, the dataset provides feature distributions, risk-ratio calculations for outcome-relevant predictor selection, and demographic distributions for age and gender initialization. The simulator was evaluated on the four most common antidepressants: fluoxetine, sertraline, trazodone, and duloxetine [53]. For the *AI Decision Aid*, the same data train prediction models that generate treatment recommendations across all 14 antidepressants. All data use complies with All of Us dissemination policies.

## Patient Simulator Design

### *Design Rationale and NIST Alignment*

The patient simulator design is grounded in the NIST AI RMF, which defines four core functions: (1) *GOVERN* - accountability and oversight, (2) *MAP* - risk identification, (3) *MEASURE* - risk assessment, and (4) *MANAGE* - risk mitigation [16]. We align with MAP and MEASURE across two complementary phases. Simulator construction supports MAP by structuring risk identification through three profile dimensions, each addressing a distinct category of risk: medical profiles targeting clinical accuracy, linguistic profiles targeting communication-dependent risks, and behavioral profiles targeting interaction-dependent risks. Simulator-AI interaction supports MEASURE by surfacing how AI performance varies across that structured profile space, facilitating downstream risk characterization.

### *Medical Profile Requirements*

Medical profiles provide the clinical foundation for risk assessment, requiring realistic and diverse medical contexts to evaluate performance across clinical heterogeneity. Generating such profiles from structured EHR data faces interconnected challenges: high-dimensional feature spaces ($10^5$-$10^6$ concept codes) with sparse task-relevant signals, data quality issues (missingness, inconsistent coding), and medically implausible feature combinations [54–56]. Our design adheres to five AI RMF-aligned trustworthiness requirements: (1) *Controllability:* emphasize task-relevant features while maintaining outcome diversity. (2) *Coherence:* maintain clinical plausibility across diagnoses, treatments, and temporal events, including rare but valid scenarios. (3) *Variability:* capture heterogeneous comorbidities and contextual factors to expose brittleness and subgroup bias. (4) *Efficiency:* balance clinical completeness with computational tractability for large-scale evaluation. (5) *Transparency:* maintain traceable feature lineage to source distributions for targeted risk analysis. The mapping of these requirements to data challenges and mitigated risks is provided in Multimedia Appendix 1.

### *Linguistic and Behavioral Profiles Requirements*

Medical profiles establish clinical context but cannot capture communication-dependent and interaction-dependent risks, requiring two additional profile dimensions, consistent with the safety taxonomy proposed by Lim et al. [57]. These are needed because accurate medical retrieval can still fail due to how a patient communicates or how the conversation unfolds. The two dimensions map onto Lim et al.'s patient input types and hazardous scenarios, informing the communication-dependent and interaction-dependent categories used here.

Communication-dependent risks emerge from heterogeneity in patient expression. Agents must maintain safety and comprehensibility across health literacy levels, condition-specific patterns, and vernacular variations [58]. Linguistic profiles enable controlled assessment grounded in health literacy and psycholinguistic research [59].

Interaction-dependent risks emerge when patients manipulate information, test boundaries, or withhold details [60]. Because evaluation cannot assess safety mechanism robustness without systematic behavioral variation, behavioral profiles are grounded in clinical and human-computer interaction research to probe agent resilience across conditions.

### *Implementation*

The patient simulator integrates three profiles: medical profiles grounded in EHR data, linguistic profiles capturing health literacy and condition-specific variation, and behavioral profiles representing engagement and interaction patterns. Profiles are operationalized through structured LLM prompting, where the medical profile assigns hierarchical indices to each clinical fact (e.g., `[3.2] Individual Psychotherapy`). For each intake question, the model identifies relevant indexed facts, applies linguistic style transfer (e.g., `[3.2]` → "talked to someone"), and constructs a JSON response containing the original facts, style-transferred equivalents, and a final natural language response with inline span markers. Simulator-provided markers are removed before being passed to the *AI Decision Aid* to ensure independent extraction. The complete prompt is provided in Multimedia Appendix 1.

### *Medical Profiles*

Phase 1 generates outcome-relevant, coherent, and diverse patient profiles. Phase 2 applies probabilistic selection using a binomial distribution derived from All of Us antidepressant response data, ensuring the final cohort reflects realistic population-level variability.

#### Medical Profile Generation (Phase 1)

Profile generation employs three core design principles [61,62]: (1) prioritize outcome-relevant features, (2) enforce statistical independence among selected features, and (3) inject controlled diversity from residual features, implemented through four stages (relevance filtering, demographic initialization, independence screening, and diversity expansion) as outlined in Figure 2.

*Stage 1: Top-K Filtering.* The algorithm restricts feature candidates to the top K predictors of the antidepressant response outcome, $e_0$ (K = 500), improving sample efficiency and aligning with the max-relevance component of mRMR feature selection [62].

*Stage 2: Demographic Seeding.* Each profile S is initialized with age and gender from All of Us demographic distributions.

*Stage 3: Independence-Screened Selection.* Features are added iteratively, with each candidate v screened for statistical independence from already-selected features. The risk ratio $RR(s, v)$ quantifies association between features s and v with respect to antidepressant response:

$$RR(s, v) = \frac{P(\text{response} \mid s \cap v)}{P(\text{response} \mid s)}$$

where values near 1 indicate independence, values $> 1.5$ positive association, and values $< 0.67$ negative association. Each candidate must satisfy:

$$1/1.5 < RR(s, v) \leq \text{high}, \quad \forall s \in S$$

This symmetric band excludes near-deterministic couplings ($RR > high$) and strong anticorrelations ($RR < 1/1.5$), ensuring statistically independent and clinically coherent feature combinations [61,62]. The upper threshold ($high$) was set to 7.

*Stage 4: Diversity Expansion.* After forming a coherent feature set, residual features outside the Top-K set are added if they satisfy:

$$RR(s, u) > 1.5, \quad \exists s \in S$$

This captures latent contextual variables enriching patient heterogeneity without compromising coherence [61,62]. The complete pseudocode is provided in Multimedia Appendix 2.

### Probabilistic Patient Selection (Phase 2)

Phase 2 selects patients whose response probabilities reproduce the All of Us population-level distribution by dividing the probability range into seven sigma-bands of a binomial distribution $B(n, p)$, with sampling weighted to match expected binomial frequencies. For example, with $n = 100$ and $p = 0.4$ ($\mu = 40$, $\sigma \approx 4.9$), approximately 64% of patients fall within $(\mu - \sigma, \mu + \sigma]$, 15% in each adjacent band $(\mu \pm 1\sigma$ to $\mu \pm 2\sigma]$, and $< 3\%$ in the tails beyond $\mu \pm 2\sigma$. This stratified sampling preserves central tendency and natural variability, while avoiding over or under representation of extreme responders. Detailed sigma-band allocations appear in Multimedia Appendix 2.

For AI RMF alignment, *Controllability* and *Coherence* are achieved through outcome-relevant feature selection and risk-ratio screening, captured by the proportion of outcome-related features and the average risk ratio. *Variability* is assessed via the number of unique concept codes, *Efficiency* by the average features per profile, and *Transparency* through explicit feature lineage enabling auditable profile construction.

```
Algorithm: Medical Profile Generation

Input: Target outcome e0, number of patients N, top-K=500

For each patient i = 1 to N:

    Stage 1: Top-K Filtering for Outcome Relevance
    V ← Top-K predictors of antidepressant response e0

    Stage 2: Demographic Seeding Initialize
    S ← {age, gender} sampled from population distributions

    Stage 3: Independence-Screened Selection
    For each candidate v in V:
        If ∀s ∈ S: 1/1.5 < RR(s,v) ≤ high:
            Add v to S

    Stage 4: Diversity Expansion
    D ← number of expanded features
    R ← residual features (outside top-K)
    count ← 0
    For candidate u in R while count < D:
        If ∃s ∈ S: RR(s,u) > 1.5:
            Add u to S
            count ← count + 1
    Compute p ← predicted response probability for profile S

Output: N patient profiles with associated response probabilities
```

Figure 2. Abbreviated medical profile generation algorithm. Complete algorithm details in Multimedia Appendix 2.

The framework has two tunable hyperparameters governing profile density: the high threshold from Stage 3 and the number of residual features added in Stage 4, set here to 3-5 per profile. Lowering high narrows the set of candidate concepts eligible for inclusion. Varying it across {4,5,6} yielded 5-8 features per profile on average, versus 8-9 at high=7, a gradual rather than sharp change indicating limited sensitivity to the exact threshold within this range. Either hyperparameter can be adjusted to target a desired profile density.

### *Linguistic Profiles*

Conversational agents must maintain safety and comprehensibility across diverse patient expression styles [63,64]; systematic linguistic variation exposes blind spots and failure modes [65,66]. The simulator implements a dual-axis linguistic framework with profiles along two independent dimensions: (1) a health literacy gradient capturing variation in comprehension, terminology, and discourse structure [19,59,67,68] and (2) condition-specific communication reflecting linguistic patterns characteristic of depression and anxiety disorders derived from Linguistic Inquiry and Word Count (LIWC)-based clinical analyses [69]. Health literacy generalizes across clinical tasks as comprehension barriers affect patient-agent interaction regardless of medical condition. Condition-specific profiles capture diagnostic patterns (here *depression* and *anxiety*) adaptable to other conditions. Table 1 specifies the five linguistic profiles.

Table 1: Linguistic user profiles across multiple dimensions.

| Profile | Key Linguistic Attributes | Example Response |
|---|---|---|
| Health Literacy | | |
| Limited | *Style:* Concrete, informal, sometimes vague.<br>*Tone:* Hesitant, uncertain, conversational.<br>*Vocab:* Everyday terms, slang, vague quantities.<br>*Structure:* Short, fragmented sentences; frequent fillers.<br>*Patterns:* Minimal elaboration unless prompted. | *"Uh, just my morning pill… you know, the one for my nerves."* |
| Functional | *Style:* Clear, basic descriptions of symptoms or routines.<br>*Tone:* Cooperative, open.<br>*Vocab:* Mix of common and medical terms.<br>*Structure*: Simple narratives; occasional causal reasoning.<br>*Patterns*: Provides coherent answers; asks clarifying questions. | *"I take Prozac every morning. It helps my mood, but I still have trouble sleeping."* |
| Proficient | *Style:* Precise, clinical, well-organized.<br>*Tone:* Confident, analytical.<br>*Vocab:* Technical terms; qualifiers such as "likely" or "seems improved."<br>*Structure*: Multi-clause, logically sequenced sentences.<br>*Patterns*: References timelines; anticipates follow-up questions. | *"I'm on fluoxetine, 20 milligrams daily. It's effective, though I've noticed mild insomnia."* |
| Condition-specific | | |
| Depression | *Style:* Brief, muted, sometimes resigned.<br>*Tone:* Flat, pessimistic, self-critical.<br>*Vocab*: Negative emotion words; self-focused phrasing.<br>*Structure*: Short, often past-tense statements.<br>*Patterns*: Withdrawn responses; dismisses reassurance. | *"Barely sleeping. My head won't shut off."* |
| Illness Anxiety Disorder | *Style:* Symptom-focused and repetitive.<br>*Tone:* Anxious, urgent.<br>*Vocab:* Symptom terms; "what if" speculation; absolutist wording.<br>*Structure:* Mix of run-on sentences and abrupt alarms.<br>*Patterns*: Reassurance-seeking cycles; future-oriented worry. | *"I felt a flutter… what if it's heart failure even though the test was normal?"* |

The linguistic profiles align with NIST requirements through literature-grounded linguistic variation [19,59,67–69], enabling controllable assessment of communication-dependent risks.

### *Behavioral Profiles*

Conversational agents struggle when patients go off-topic, respond vaguely, or withhold information [70,71]. Systematic behavioral variation is essential for stress-testing agent robustness and assessing recovery from conversational breakdowns [72,73]. We organized the 13 patient behaviors documented by Simpson et al. [74] into four behavioral categories (complete mapping in Multimedia Appendix 2), with *Structured and Cooperative* as an additional fifth category. This study operationalizes three profiles: *Distracted & Unfocused* and *Adversarial & Combative* to capture challenging dynamics, with *Structured & Cooperative* as baseline. We selected this subset to bound experimental scope while preserving the contrast that matters for risk assessment: a cooperative baseline against two profiles that diverge in distinct ways. Each profile varies along four dimensions: conversational adherence, engagement, topical focus, and adversarial behavior, as detailed in Table 2.

Table 2: Beha0vioral user profiles across multiple dimensions.

| **Profile** | **Key Attributes** | **Example Response** |
|---|---|---|
| Structured & Cooperative | *Adherence:* High | *"Yes, I take 20 mg of fluoxetine every morning around 8 AM. I haven't missed a dose in the last three weeks."* |
| | *Engagement:* High | |
| | *Topical Focus:* High | |
| | *Adversarial / Toxic Behavior:* Minimal | |
| Distracted & Unfocused | *Adherence:* Low | *"I was… wait, which one? Oh right, yeah I think? But yesterday I forgot — also my dog wouldn't eat."* |
| | *Engagement:* Sporadic | |
| | *Topical Focus:* Off-topic | |
| | *Adversarial / Toxic Behavior:* Inadvertent derailment of conversation | |
| Adversarial & Combative | *Adherence:* Variable | *"What kind of dumb question is that? Maybe if your system worked better, I wouldn't have to answer this again."* |
| | *Engagement:* Variable | |
| | *Topical Focus:* Variable | |
| | Adversarial / Toxic Behavior: Overtly confrontational or hostile | |

The behavioral profiles align with NIST requirements through empirically grounded behavioral variation [74], enabling controlled assessment of interaction-dependent risks including adversarial scenarios with transparent lineage.

Together, these three profile dimensions provide comprehensive MAP and MEASURE coverage for conversational healthcare AI risk assessment.

### Chain-of-Thought Prompting Strategy

We employ Chain-of-thought (CoT) prompting [75] to generate realistic patient responses through a structured process. For each question posed by the *AI Decision Aid*, the prompt directs the model to (1) identify relevant indexed medical attributes, (2) apply controlled term-level linguistic transformations, and (3) construct a natural language response with explicit references. Behavioral constraints are enforced as rules governing interaction, while linguistic constraints shape the form of expression. All outputs use a fixed JSON schema, enabling traceability and systematic evaluation (details in Multimedia Appendix 1).

### AI Decision Aid

The *AI Decision Aid* is a multi-agent conversational platform for antidepressant selection, serving as the system under evaluation in this black-box assessment. The system conducts structured intake through six sequential stages: establishing rapport, collecting illness history, gathering antidepressant history, documenting current medications, recording clinical procedures, and generating personalized recommendations. Each stage employs LLM-guided dialogue to elicit clinical information and a Retrieval-Augmented Generation (RAG) system to normalize medical concepts before passing them to an analytical reasoning system for estimating antidepressant response. Concept normalization is a two-step process: an embedding-based retriever returns the top-K candidate concepts from the study lexicon for each patient utterance, and an LLM selects the single best match from those candidates. Only this rank-1 selection is forwarded to the analytical model, and downstream candidates are discarded. If the correct concept is in the top-K but not ranked first, it does not reach the recommendation step. The analytical advice system is based on the Direct Effects Multiplicative Inference Algorithm (DEMI) [76], which estimates dependent Bayesian relationships among medical concepts to support clinical reasoning under partial observability. Alternative recommendation models (e.g., logistic regression) are compatible. System specifications and code are publicly available [24].

### Evaluation Framework

The evaluation framework examines patient simulator performance across medical, linguistic, and behavioral dimensions: medical profiles through human and LLM-based annotation using a three-label schema (Accurate, Inaccurate, Unsupported); linguistic and behavioral profiles through human annotation, quantitative metrics, and visual clustering.

#### *Human Annotation*

All annotations were performed by two annotators with complementary domain expertise: one with an undergraduate degree in Psychology and one a Registered

Nurse. Each sample was independently annotated by both raters, with disagreements resolved through adjudication.

### *Medical Profile Evaluation*

The patient simulator CoT process: (1) identifies relevant medical concepts, (2) rephrases each concept according to the linguistic profile, and (3) constructs a natural language response based on linguistic and behavioral characteristics. For each conversational turn, the simulator outputs (a) relevant concepts referenced numerically (e.g., [2.3]) and (b) a natural language response with inline spans marking where each concept is expressed. These spans serve as annotation units labeled using a three-category schema: ***Accurate***, where the medical fact is correctly expressed with minor colloquialisms (e.g., “happy pills” for “Prozac”) permitted if core clinical meaning is preserved; ***Inaccurate***, where the fact is present but misrepresented, distorting critical clinical details or using implausible phrasing; and ***Unsupported***, where content does not correspond to the patient’s profile, capturing hallucinated or fabricated details outside tagged spans. This approach evaluates expression fidelity conditioned on correct concept retrieval. Concept recall is computed programmatically from the simulator’s traced outputs, with 95% of profile concepts appearing at least once in the generated conversations.

We validate the annotation and judging pipeline itself through controlled error injection. The patient simulator expresses nearly all medical concepts accurately, so annotators could achieve artificially high inter-annotator agreement (IAA) by labeling all concepts as *Accurate*, and the LLM judge validation would lack discriminative power. To enable rigorous evaluation, we introduce controlled semantic perturbations by replacing clinical concepts with semantically similar but clinically distinct alternatives (e.g., “hypertension” → “prehypertension”, “diabetes mellitus” → “prediabetes”), selected via semantic search over SNOMED CT and Current Procedural Terminology, 4th edition (CPT-4) that identifies the top 20 candidates, randomly shuffled before ontology-based filtering removes hierarchical ancestors or descendants and enforces minimum hierarchical distance. The first qualifying candidate is selected, ensuring variation rather than defaulting to the nearest semantic neighbor; if none qualifies, the threshold is relaxed or pool expanded. The resulting perturbed profiles maintain linguistic realism while introducing subtle clinical inaccuracies for evaluating both human annotation quality and LLM judge performance.

### *Linguistic Profile Evaluation*

Linguistic profiles are evaluated through human annotation, where annotators classify each conversation into one of five predefined linguistic profiles. Five automated metrics are also used:

1. **Reading Level:** Flesch-Kincaid Grade Level (FKGL) [77] estimates school grade level and is calculated per response turn and averaged per conversation.

2. **Average Response Length:** Average words per turn using NLTK's English tokenizer [78].
3. **Medical Term Density:** Clinical term count via greedy n-gram matching (up to 6-grams) against the study medical lexicon (all concept codes present in the database).
4. **Depression Score:** Mean turn-level probability from an XLM-RoBERTa-based depressive symptom classifier [79].
5. **t-SNE:** Projects linguistic features to two dimensions using cosine distance; multiple seeds tested to minimize KL divergence.

### *Behavioral Profile Evaluation*

Annotators classify each conversation into one of three behavior profiles. Three automated metrics quantify behavioral patterns:

1. **On-topic similarity:** Average cosine similarity between each *AI Decision Aid* turn and the patient simulator response.
2. **Toxicity:** Mean turn-level probability from the XLM-RoBERTa-based toxicity classifier [80].
3. **t-SNE:** Projects behavioral features analogously to linguistic t-SNE.

### *LLM Judge Evaluation*

LLM-based judges have demonstrated strong performance in healthcare evaluation tasks [81–83]. Automated NLP similarity metrics, by contrast, have not been shown to correlate with expert human evaluation of generative LLM outputs on EHR data [84]. We employ a separate LLM, *Claude Opus 4.6*, as an automated judge, applying the same annotation schema across medical, linguistic, and behavioral dimensions. The judge prompt was developed on 45 held-out conversations without human annotations. Judge alignment was then validated against human annotations on the full evaluation set, with results presented in the Results section.

## Experimental Paradigm

Experiments used 60 medical profiles evaluated across four antidepressants (Sertraline 17%, Trazodone 15%, Fluoxetine 13%, and Duloxetine 11%) reflecting common clinical practice [53]. For each antidepressant, a pool of patient profiles was simulated and down-sampled to 15 profiles per drug by stratifying across response probability bands. With *high*=7 and 3-5 residual features per profile, generation yielded an average of 8-9 clinical features per profile, consistent with reports that outpatient visits address 5.4–7.1 clinical items per encounter [85]. This feature density reflects plausible patient disclosure and contrasts with All of Us records (approximately 98.88 features per patient).

The design comprises three settings: (1) five linguistic profiles with fixed *Structured & Cooperative* behavior (300 conversations), (2) three behavioral profiles with fixed *Functional Health Literacy* (180 conversations), and (3) combined linguistic-behavioral variation (150 conversations), yielding 500 unique simulated conversations. The patient simulator and *AI Decision Aid* are powered by *GPT-4.1*,

*Claude Opus 4.6* serves as the LLM judge, and all embedding-based analyses including t-SNE projections and on-topic similarity use OpenAI's text-embedding-3-small, which also supports RAG-based concept normalization in the *AI Decision Aid*. Both are implemented as separate agents within LangGraph [86].

### Ethical Considerations

This project was examined by George Mason University's Institutional Review Board (study 2154028-1 for work on the All of Us database and study 00000437 for evaluation of text of the advice through annotation) and was determined to be "not research" in the context of the definition of research on human subjects by the United States Department of Health and Human Services. Informed consent was therefore not required. All of Us data were accessed exclusively through the program's secure Researcher Workbench under the data use agreement, and no individual-level records are redistributed. Annotation work was performed by named research personnel acknowledged in the manuscript. No human subjects were recruited for this study.

## Results

### Simulated Medical Profile Validation

Across 60 medical profiles, the framework demonstrates systematic risk characterization aligned with the NIST AI RMF MAP and MEASURE functions. *Controllability* was achieved by emphasizing outcome-related features (37.2% of features per profile) while maintaining outcome diversity across response probability bands. *Coherence* was reflected in an average risk ratio of 2.64 among outcome-related features, consistent with the risk-ratio gating. *Variability* was maintained through 292 unique concept codes across the 60 profiles. *Efficiency* was demonstrated by an average of 8.08 features per profile, substantially lower than the 98.88 features in a typical All of Us record. *Transparency* was maintained through explicit feature provenance across all generation components: demographic attributes followed All of Us distributions (ages 13-19: 0.1%, 20-40: 25.3%, 41-64: 41.5%, 65-79: 28%, 80-89: 5.1%; male: 37.3%, female: 62.7%); outcome relevance was enforced using the Top-K set; residual features were drawn from the non-Top-K pool; and outcome probabilities matched antidepressant-specific binomial distributions: fluoxetine (0.28-0.57, mean 0.41), sertraline (0.30-0.59, mean 0.44), duloxetine (0.28-0.57, mean 0.43), and trazodone (0.16-0.42, mean 0.29).

### Medical Profile Validation

Medical profile fidelity was assessed through 1,786 simulator-generated medical concepts with annotated spans across 100 conversations, including 292 perturbed concepts. Table 3 shows IAA between human annotators and the LLM judge. Across all concepts, human IAA was high (0.73 $\kappa$, 0.93 F1). Agreement decreased for perturbed concepts (0.30 $\kappa$, 0.76 F1), reflecting the difficulty of subtle semantic substitutions. Human-LLM judge agreement followed the same pattern: 0.78 $\kappa$, 0.93 F1 overall; 0.24 $\kappa$, 0.77 F1 on perturbed concepts. Annotators identified 109 unique

*Unsupported* cases, predominantly incidental medication mentions (e.g., Tylenol, ibuprofen, vitamins) and minor symptom references (e.g., headache, neck pain) not present in the medical profiles (IAA of 0.75 F1). The LLM judge identified 76 *Unsupported* cases of the same type (0.58 F1 against adjudicated human annotations). We also performed a paired bootstrap test [87] with 10,000 resamples. We found that human-human agreement (κ = 0.73) was slightly higher than agreement between individual human annotators and the LLM (κ = 0.70-0.71), but these differences were not statistically significant ($P$=.21, $P$=.32). Agreement between the LLM and each annotator also did not differ significantly ($P$=.63), indicating no systematic annotator-specific bias. Overall, LLM–human agreement was comparable to human–human agreement within sampling variability.

Table 3. Agreement scores for human–human and human–LLM evaluations across full and perturbed concept sets.

| **Evaluation** | **Concepts Set** | **n** | **Cohen's κ** | **Accurate F1** | **Inaccurate F1** | **Unsupported F1** | **Overall F1** |
|---|---|---|---|---|---|---|---|
| Human-Human | Perturbed | 292 | 0.30 | 0.45 | 0.85 | - | 0.76 |
| | Full | 1786 | 0.73 | 0.96 | 0.76 | 0.75 | 0.93 |
| Human-LLM | Perturbed | 292 | 0.24 | 0.37 | 0.87 | - | 0.77 |
| | Full | 1786 | 0.78 | 0.97 | 0.81 | 0.58 | 0.93 |

In the LLM judge evaluation of medical profiles, 96.6% of 8,210 expressed concepts were labeled *Accurate*, 1.1% *Inaccurate*, and 2.3% *Unsupported*. The simulator maintained high accuracy (96-99%) and low error rates across all profiles. Within this range, concept expression frequency increases across the literacy gradient, peaking with *Proficient*. Error types diverge modestly by profile: *Proficient* shows the highest *Unsupported* rate (1.0%) due to greater medical term density, while *Illness Anxiety Disorder* produces the highest inaccuracy rate (2.5%) due to its repetitive tone. The *Depression* profile produces the fewest expressed concepts with minimal errors.

Table 4 breaks down fidelity across linguistic and behavioral profiles. *Structured & Cooperative* ensures stability, *Distracted & Unfocused* introduces frequent *Unsupported* outputs, and *Adversarial & Combative* minimizes errors as confrontational responses tend to be terse, limiting opportunities for errors.

Table 4: LLM Judge Evaluation of Medical Profile fidelity across linguistic and behavioral profiles.

| **Profile Name \ Metrics** | | **Total** | **Accurate** | **Inaccurate** | **Unsupported** |
|---|---|---|---|---|---|
| **Linguistic profiles under *Structured & Cooperative* behavioral condition.** | | | | | |
| Health Literacy | Limited | 863 | 98.4% | 1.3% | 0.3% |
| | Functional | 952 | 98.2% | 1.4% | 0.4% |
| | Proficient | 981 | 98.7% | 0.3% | 1.0% |
| Condition-specific | Depression | 842 | 98.1% | 1.1% | 0.8% |
| | Illness Anxiety Disorder | 1,192 | 96.1% | 2.5% | 1.3% |
| | **Subtotal** | **4,830** | **97.8%** | **1.4%** | **0.8%** |
| **Behavioral profiles under *Functional Health Literacy* linguistic condition.** | | | | | |
| Behavioral | Structured & Cooperative | 952 | 98.2% | 1.4% | 0.4% |
| | Distracted & Unfocused | 988 | 89.5% | 0.5% | 10.0% |
| | Adversarial & Combative | 1,095 | 99.3% | 0.0% | 0.7% |
| | **Subtotal** | **3,035** | **95.7%** | **0.6%** | **3.7%** |

The intersection of linguistic and behavioral profiles shapes error type rather than overall accuracy, which remains high across all combinations (Figure 3). *Distracted & Unfocused* produces elevated *Unsupported* rates in lower health literacy profiles, where conversational drift introduces off-profile content; *Adversarial & Combative* drives inaccuracies in *Proficient* and *Illness Anxiety Disorder* profiles, where confrontational responses distort rather than fabricate clinical content; and *Structured & Cooperative* yields the lowest error rates, though *Illness Anxiety Disorder* produces *Unsupported* content even under cooperative conditions.

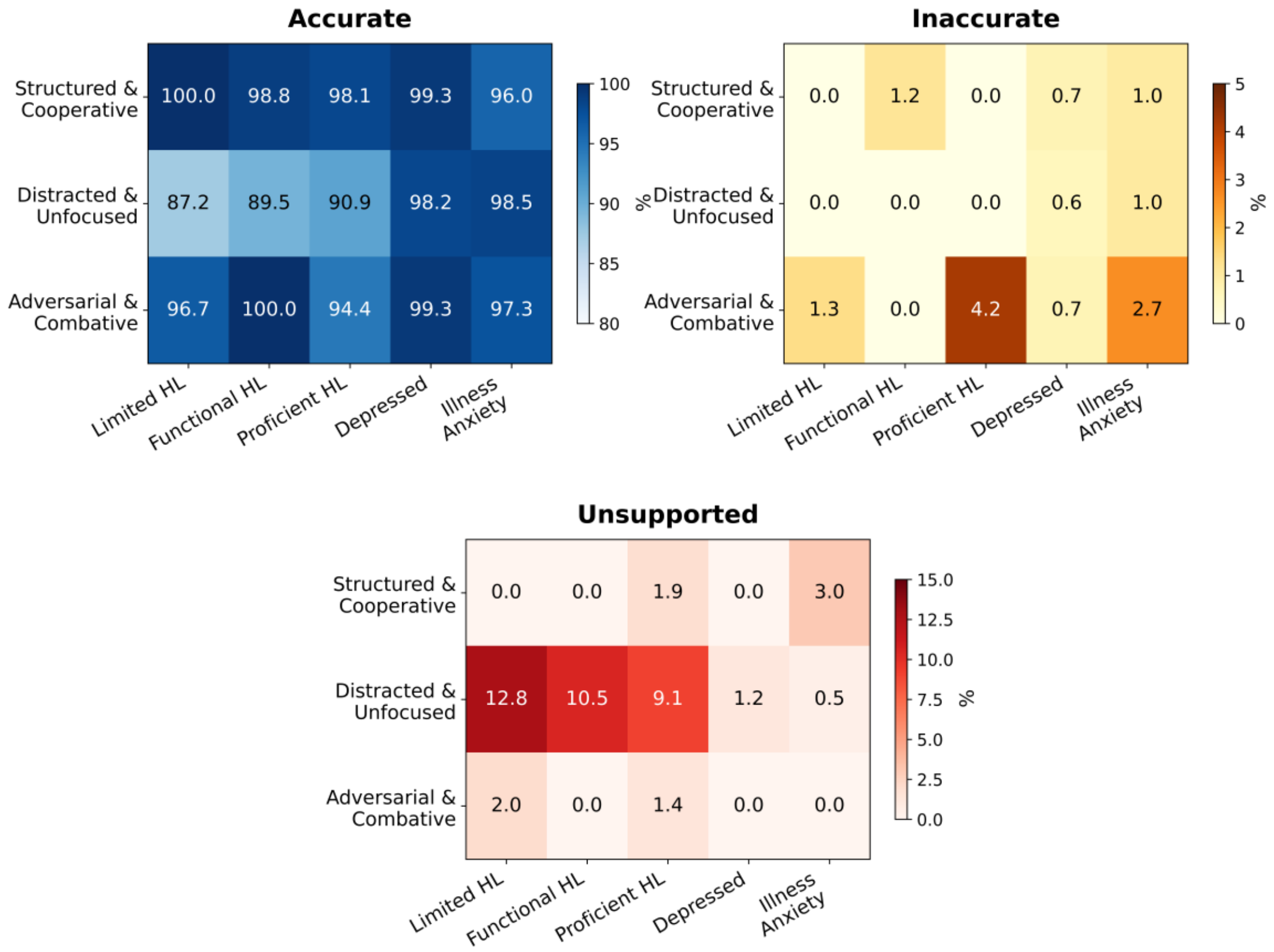

Figure 3: LLM judge evaluations from the intersection of linguistic and behavioral profiles.

## Linguistic Profile Validation

Human annotators showed substantial agreement on linguistic profile classification (0.61 $\kappa$, 0.70 micro F1), comparable to Human-LLM agreement (0.63 $\kappa$, 0.70 micro F1). Adjudicated labels against predefined profiles reached 0.87 micro F1, indicating consistent expression of intended profiles, and the LLM judge achieved 0.95 micro F1 across 500 conversations.

Table 5 shows linguistic variation under fixed *Structured & Cooperative* behavior. Reading level increases monotonically from *Limited* (3.59 FKGL) to *Proficient* (11.90 FKGL), with response length and medical term density following the same gradient (*Limited*: 34.34 words, 4.33 medical terms; *Proficient*: 38.91 words, 11.68 medical terms). The *Depression* profile produces the shortest responses (21.48 words, depression score 0.20), whereas *Illness Anxiety Disorder* produces the longest (65.05 words, medical term density 14.27, depression score 0.25), reflecting greater health-related elaboration.

Table 5: Evaluation of linguistic profiles under a *Structured & Cooperative* behavioral condition.

| **Profile Name \ Metrics** | | **Reading Level (FKGL)** | **Response Length** | **Medical Term Density** | **Depression Score** |
|---|---|---|---|---|---|
| Health Literacy | Limited | 3.59 | 34.34 | 4.33 | 0.01 |
| | Functional | 7.63 | 28.70 | 9.83 | 0.00 |
| | Proficient | 11.90 | 38.91 | 11.68 | 0.00 |
| Condition -specific | Depression | 4.49 | 21.48 | 6.17 | 0.20 |
| | Illness Anxiety Disorder | 8.11 | 65.05 | 14.27 | 0.25 |

Figure 4 shows t-SNE clustering where linguistic profiles form distinct clusters, with *Functional* and *Proficient* overlapping due to their shared characteristics and *Depression* partially overlapping with lower literacy profiles, while *Illness Anxiety Disorder* forms a distinct cluster. Together, these confirm graded, distinct linguistic profiles.

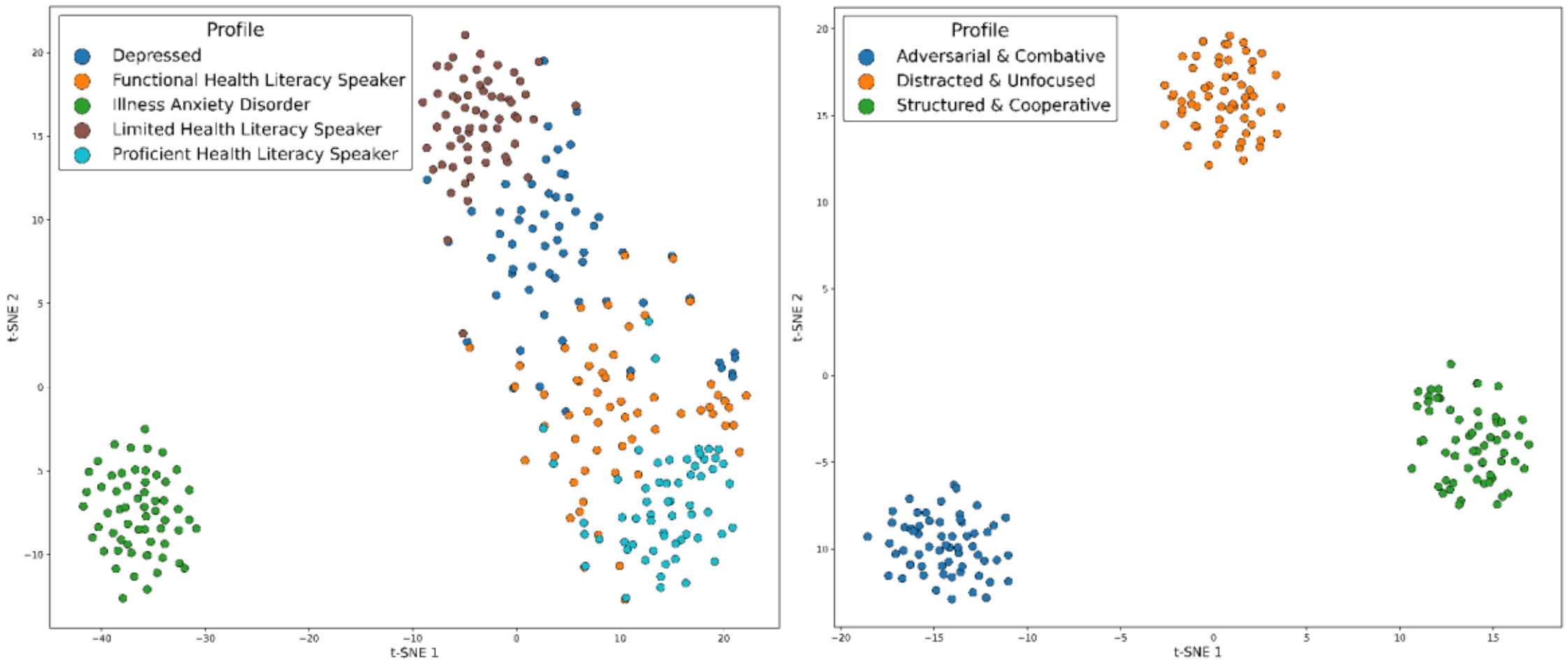

Figure 4: t-SNE visualization of response embeddings for linguistic profiles (left, under *Structured & Cooperative* behavioral condition) and behavioral profiles (right, under *Functional Health Literacy* linguistic condition).

## Behavioral Profile Validation

Human annotators showed high agreement on behavioral profile classification (0.93 $\kappa$, 0.96 micro F1), identical to Human-LLM agreement, with classification against predefined profiles reaching 0.98 micro F1. Under fixed *Functional Health Literacy* (Table 6), *Structured & Cooperative* achieves baseline on-topic similarity (0.51) and lowest toxicity (0.0003). *Distracted & Unfocused* shows slightly reduced on-topic similarity (0.50), consistent with conversational drift, while maintaining low toxicity (0.0035). *Adversarial & Combative* exhibits substantially higher toxicity (0.0564) while retaining the highest topical relevance (0.52).

Table 6. Evaluation of behavioral profiles under a *Functional Health Literacy* linguistic condition.

| **Profile** | **On-topic Similarity** | **Toxicity** |
|---|---|---|
| Structured & Cooperative | 0.51 | 0.0003 |
| Distracted & Unfocused | 0.50 | 0.0035 |
| Adversarial & Combative | 0.52 | 0.0564 |

Figure 4 shows t-SNE clustering with clear separation among behavioral profiles. Together, these confirm distinct behavioral profiles under fixed linguistic conditions.

### Profile Interactions

Figure 5 presents intersectional evaluation across linguistic and behavioral profile combinations. *Distracted & Unfocused* produces the longest responses across all linguistic profiles (61-82 words vs. 21-66 for *Structured & Cooperative*), while *Depression* consistently produces the shortest (21-32 words) with elevated depression scores (0.18-0.25). Toxicity shows the strongest behavioral dominance, with *Adversarial & Combative* exhibiting elevated toxicity (0.03-0.11) regardless of linguistic profile, whereas reading level and depression scores remain largely determined by linguistic profiles. On-topic similarity shows modest behavioral effects, with *Structured & Cooperative* achieving slightly higher alignment than *Distracted & Unfocused*. These patterns demonstrate that the simulator maintains profile independence where intended (linguistic features preserved across behaviors) while capturing realistic interactions (behavioral effects on engagement and tone).

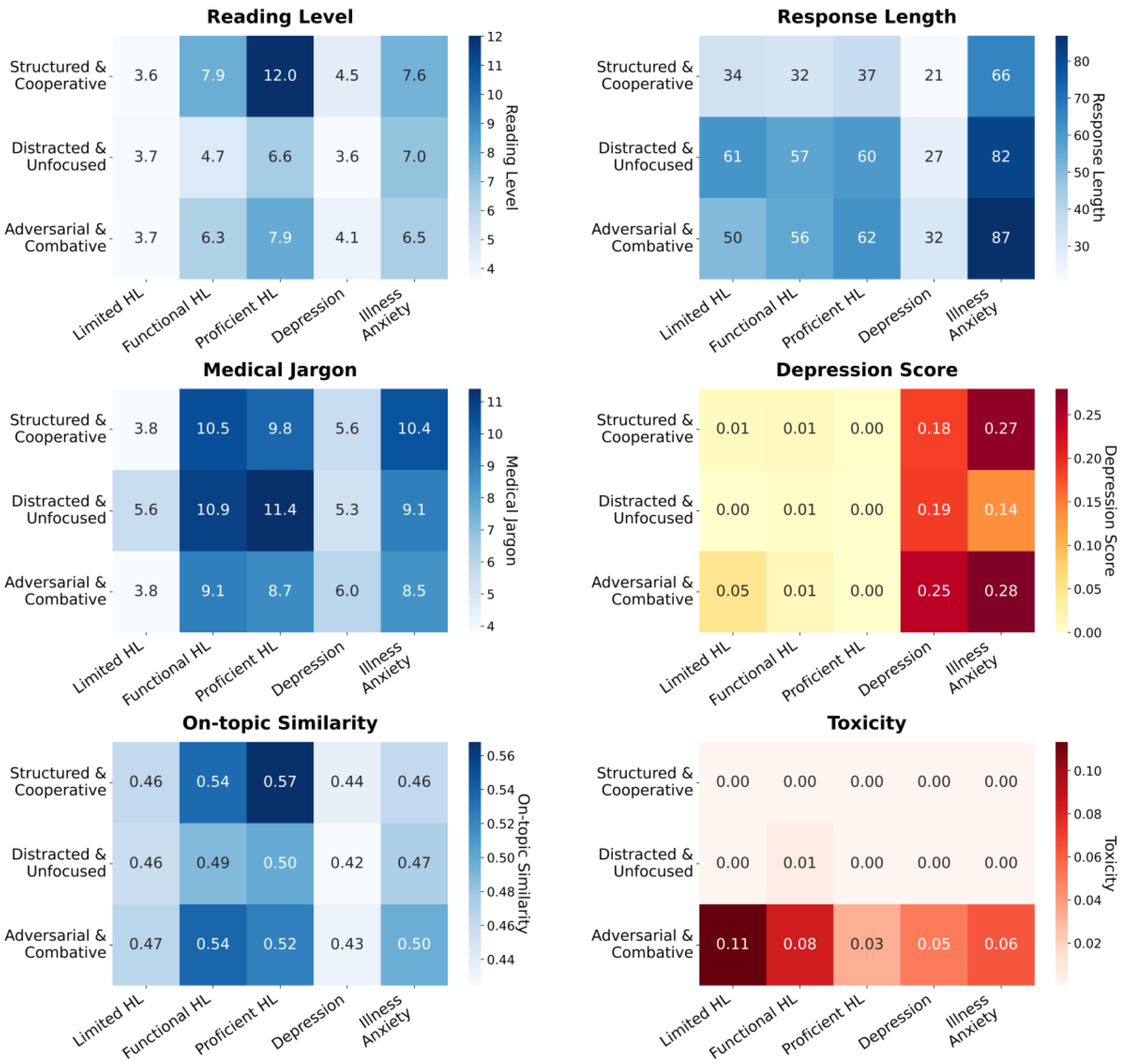

Figure 5: Intersection of linguistic and behavioral profiles across evaluation metrics.

## AI Decision Aid Performance: Risk Measurement Across Patient Variation

This section evaluates *AI Decision Aid* performance across 500 simulated conversations.

### *Concept Retrieval Coverage*

Overall concept recall reached 93.0% across 2,819 reference concepts, indicating approximately 7% were not identified by the *AI Decision Aid*. Recall was highest for diagnoses (94.3%, $n = 2{,}077$) and procedures (91.0%, $n = 569$), while medications showed the lowest recall (84.4%, $n = 173$), identifying medication retrieval as a relative vulnerability. The system additionally introduced 151 concepts outside predefined profiles, predominantly medications (84) and procedures (57), reflecting conversational elaboration.

### *Retrieval Accuracy and Health Literacy Effects*

The retrieval performance of concepts identified by the *AI Decision Aid* is reported on the full set of simulator-generated reference concepts (n = 2,819), with unidentified concepts contributing to overall outcomes. Overall rank-1 accuracy was 65.9%; however, 80.2% of concepts appeared within the top 20 candidates, indicating ranking imprecision rather than missing semantic coverage. Retrieval accuracy increases with health literacy level. Rank-1 accuracy ranges from 47.6% (*Limited*) to 81.9% (*Proficient*). Condition-specific profiles fall between: *Depression* at 62.0% and *Illness Anxiety Disorder* at 63.7%. Vague or affect-heavy language reduces ranking accuracy (Multimedia Appendix 3).

Patients with *Limited* health literacy or *Depression* use colloquial, fragmented, or affect-heavy language that misaligns with the system's canonical concept vocabulary. Because Top-20 retrieval does not fully close this gap, the system introduces a structural bias: retrieval is reliable for patients who use precise medical terminology but degrades for populations expressing equivalent clinical content through less structured language (see Multimedia Appendix 3).

### *Downstream Risk to Antidepressant Recommendation*

Recommendations are produced by DEMI over 15 antidepressants plus a No Recommendation category (probability < 0.1), making downstream performance sensitive to information loss during intake. Performance improves with higher health literacy, with *Proficient* achieving the highest F1 (0.73) under *Structured & Cooperative* and *Limited* the lowest (0.48 F1). Behavioral profiles further modulate this risk: *Structured & Cooperative* yields more stable outcomes, while *Distracted & Unfocused* and *Adversarial & Combative* are each associated with performance shifts that vary by linguistic profile as shown in Figure 6.

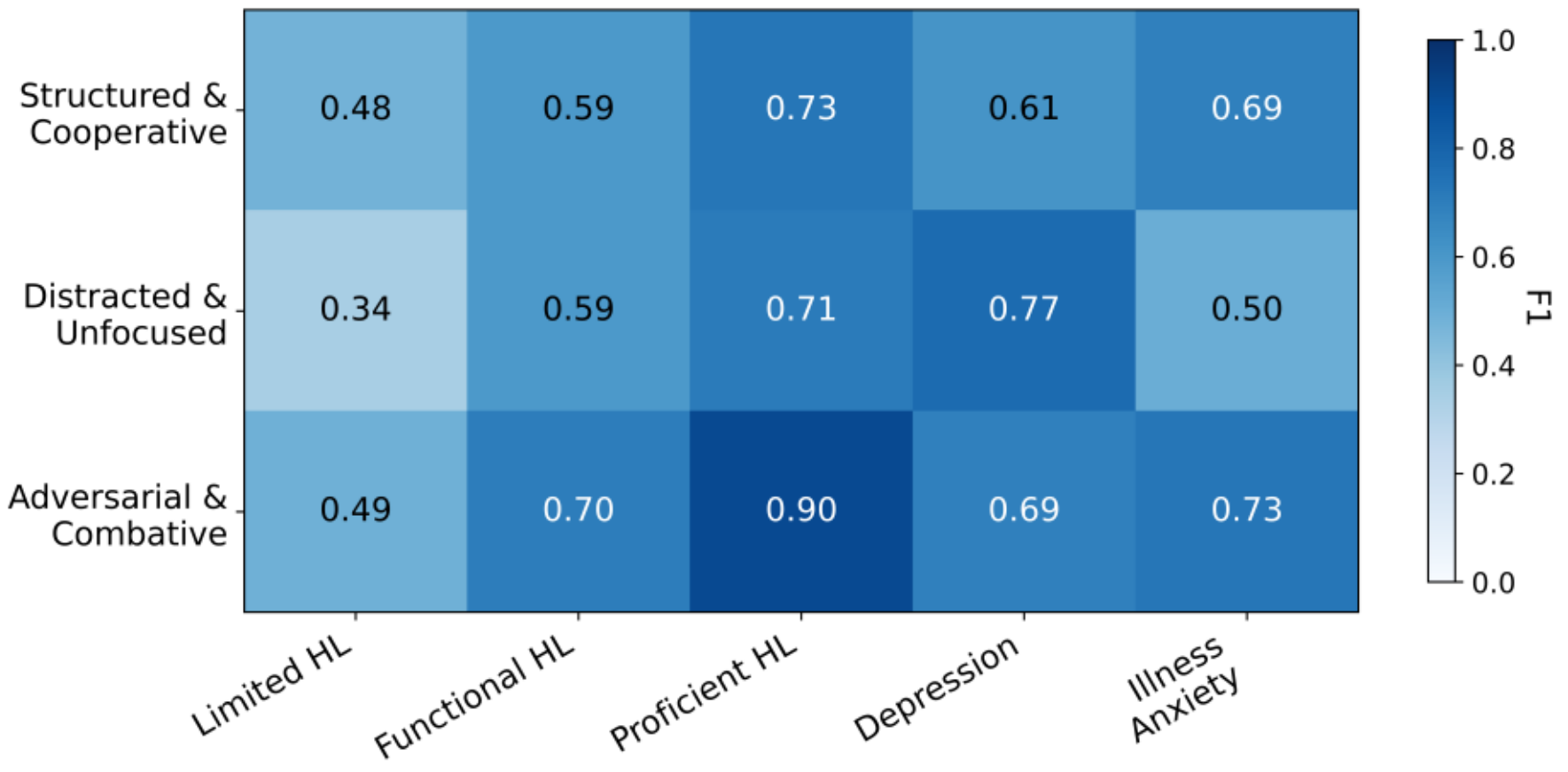

Figure 6: F1 Scores for antidepressant recommendation before and after *AI Decision Aid* processing across linguistic and behavioral profile combinations.

## Discussion

This study demonstrates that patient simulation, grounded in the NIST AI RMF, can systematically expose risks in conversational healthcare AI that static evaluation methods miss. Across 500 simulated conversations, the framework revealed a monotonic performance gradient across health literacy levels, with corresponding effects on antidepressant recommendation accuracy. The remainder of this section discusses these principal findings, their clinical implications, and the limitations of the work.

### Principal Findings

#### *Medical Profile Generation*

The medical profile generation process produced clinically coherent profiles with explicit feature lineage traceable to source EHR data. To validate both annotator and LLM judge performance, we applied controlled error injection. Semantically similar but clinically distinct perturbations (e.g., hypertension → prehypertension, diabetes mellitus → prediabetes, major depressive disorder moderate → major depressive disorder mild, generalized anxiety disorder → adjustment disorder with anxiety) tested annotator and LLM judge sensitivity. The injection is hard by design. Each perturbation is semantically close to the original concept, the kind of subtle substitution an LLM is likely to produce, and once rendered in a patient's communication style it is easily mistaken for an accurate expression. Agreement on perturbed concepts was lower than overall agreement (0.30 κ vs. 0.73 κ human-human; 0.24 κ vs. 0.78 κ human-LLM). This partly reflects the intended difficulty of the perturbations, which approximate real-world concept ambiguity, but also signals a real detection limit relevant to deployment monitoring. If human raters and the LLM judge both struggle to catch subtle clinical substitutions under these controlled conditions, comparable errors are unlikely to be easier to catch in live use, where similar ambiguity could go undetected without dedicated monitoring.

#### *Linguistic and Behavioral Profile Effectiveness*

The simulator produced measurably distinct profiles across both dimensions. Behavioral profiles formed categorically different clusters, while linguistic profiles formed a continuous health literacy gradient; the *Functional* and *Proficient* overlap reflects their genuine similarity along this continuum. Cross-dimensional analysis confirmed profile independence with two interaction effects: *Distracted & Unfocused* consistently increased response length, and *Adversarial & Combative* increased toxicity regardless of linguistic profile.

#### *Health Literacy as Risk Factor for Clinical AI*

Linguistic profile substantially affected *AI Decision Aid* performance with monotonic degradation across the health literacy gradient: rank-1 retrieval ranged from 47.6% (*Limited*) to 69.6% (*Functional*) to 81.9% (*Proficient*), with the gradient persisting in downstream antidepressant recommendation accuracy. The system relies on precise terminology: colloquial expressions like "my morning pill for my nerves" force the retrieval system to resolve a larger semantic distance than "20 mg of fluoxetine for generalized anxiety disorder." These failures point to specific *MANAGE*-function interventions: clarification prompting, terminology normalization, or multi-pass retrieval. Condition-specific profiles showed intermediate retrieval performance (*Depression* 62.0%, *Illness Anxiety Disorder* 63.7%), suggesting affective expression also degrades performance, though less severely than low health literacy.

### *LLM Judge Viability*

The LLM judge aligned strongly with human annotators on medical concept agreement (0.93 F1 overall) and showed lower agreement on perturbed concepts (0.77 F1), mirroring divergence among human annotators on those same items. The judge also flagged *Unsupported* content beyond the perturbations, related to incidental medications and minor symptoms not present in the profiles. These results support LLM judge use for large-scale screening, with human review reserved for edge cases.

## Comparison with Prior Work

Patient simulators have advanced rapidly, yet none jointly addresses medical, linguistic, and behavioral risk within a single risk-aligned framework. PATIENT-Ψ [47] achieves strong behavioral and emotional realism through 106 expert-authored CBT cognitive schemas, with a focus on clinical training rather than EHR-grounded risk assessment. PatientSim [35] benchmarks LLM doctor agents using MIMIC-derived clinical profiles and a four-axis persona space (personality, language proficiency, recall, confusion), with persona treated as a fixed categorical attribute and traceability provided at the persona level rather than the concept level. MATRIX [57] audits hazards across 2,100 dialogues using a safety-engineered hazard taxonomy, a simulated patient (PatBot), and an LLM judge, with clinical content drawn from scenario templates and a self-contained safety taxonomy rather than alignment to an external risk-management framework. Cook et al. [34] demonstrate training scalability through prompt-engineered GPT virtual patients on two outpatient topics, with the evaluation directed at clinician history-taking rather than at the AI system itself. In contrast, our patient simulator unifies three risk-aligned profile dimensions in a single architecture: EHR-grounded medical profiles drawn from All of Us via risk-ratio gating, five linguistic profiles spanning three health literacy levels (*Limited*, *Functional*, *Proficient*) and two condition-specific styles (*Depression*, *Illness Anxiety Disorder*), and three behavioral profiles (*Structured & Cooperative*, *Distracted & Unfocused*, *Adversarial & Combative*). The framework aligns with the NIST AI RMF MAP and MEASURE functions and provides concept-level lineage from each utterance back to source clinical facts.

## Clinical Implications

Performance differences across health literacy and behavioral profiles raise clinical, equity, and governance concerns that should be addressed before the *AI Decision Aid* is deployed.

### *Health Literacy as a Patient Safety Issue*

Rank-1 retrieval falls from 81.9% (*Proficient*) to 47.6% (*Limited*). Because antidepressant trials typically last six to twelve weeks before effectiveness can be assessed [88,89], retrieval errors may not surface until after weeks of ineffective or harmful treatment. *Limited health literacy* is more common among patients with severe depression, lower socioeconomic status, advanced age, and less formal education [90,91], the same groups that bear the heaviest burden of untreated mental illness [92,93]. The performance gap is concentrated in the population most in need of accessible support. This is not a technical limitation to be addressed in a later release; it is an immediate patient safety risk requiring active mitigation before deployment.

### *Behavioral Profiles Reflect Psychiatric Phenotypes*

The *Distracted & Unfocused* profile mirrors the psychomotor slowing, impaired attention, and disorganized thought of moderate-to-severe depression [94,95], core symptoms of the illness rather than mere variations in communication style. The *Adversarial & Combative* profile often reflects trauma history, prior negative healthcare experiences, or treatment frustration rather than a difficult personality [96,97]. The system is least reliable for patients whose illness severity itself makes communication difficult. These clinical interpretations explain why the observed performance gaps carry safety significance; the behavioral profiles were derived from a general interaction taxonomy [74] and have not been validated against symptom presentations aligned with the Diagnostic and Statistical Manual of Mental Disorders (DSM), so the mapping to specific psychiatric phenotypes should be read as illustrative rather than diagnostic.

### *Care Pathway Integration Determines Harm Potential*

Provider-supervised use makes the identified gaps manageable; direct-to-patient deployment, particularly to underserved populations with limited provider access, amplifies them. Deployment plans should specify the oversight model, review criteria, and escalation pathways for low-confidence recommendations. Under the NIST AI RMF, these governance structures correspond directly to MANAGE-function commitments and should be specified before deployment approval. Where the system informs treatment selection, providers should also interpret its outputs as patterns of medication persistence, recognizing that persistence and symptom remission can diverge in real-world treatment, with prognostic implications [98]. The MANAGE-function interventions above (clarification prompting, terminology normalization, adaptive retrieval) should be developed with input from patients with limited health literacy and providers experienced in psychiatric communication, and validated in the same simulation framework before rollout.

### *Equity and Participatory Design*

The performance gradient documented here is a familiar pattern in healthcare AI. Tools designed and validated on data representing dominant communication styles reinforce existing disparities at deployment scale rather than reducing them [99]. Antidepressant selection compounds this risk. Comorbidity, pharmacogenomics, and social determinants all shape outcomes, and patients with limited health literacy already face diagnostic and treatment gaps in standard psychiatric care [100,101]. A system that underperforms for them risks widening those gaps. Technical correction alone will not address this. The MANAGE-function interventions identified above should be co-designed with patients carrying limited health literacy and with clinicians who serve them. Psychiatric care has developed practices for this work, including teach-back protocols, plain language standards, and culturally adapted patient education [102]. Conversational AI for these populations should be built on the same foundation.

## Limitations

### *Generalizability*

Simulated conversations have not been validated against real patient interactions; ecological validity will be evaluated in a planned real-world trial. The evaluation covers a single clinical task and four antidepressants. The framework itself does not assume a deployment model, but the harm potential of the performance gaps we report depends on how the AI Decision Aid is deployed; future evaluations should match the oversight model under which it will be used.

### *Validity of Clinical Signals*

Antidepressant response was operationalized as ten-week medication persistence without switching or augmentation, a surrogate influenced by formulary access, provider inertia, and barriers to follow-up that does not capture clinical remission [103,104]. Future work should anchor response in patient-reported outcomes or validated symptom scales (e.g., PHQ-9, HAM-D) [105,106].

### *Profile Construction*

Medical profiles relied exclusively on structured EHR data, omitting the unstructured clinical narrative where clinicians document symptom severity, treatment history, and psychosocial context. The independence screening criterion may exclude valid coupled comorbidities, underrepresenting patients with strongly correlated conditions. Behavioral profiles follow Simpson et al.'s taxonomy [74] and are not validated against DSM-aligned presentations: the *Distracted & Unfocused* profile resembles depressive neurovegetative symptoms but is not clinically verified, and the *Adversarial & Combative* profile does not distinguish oppositional engagement from trauma- or frustration-driven presentations. Validation against psychiatrically characterized populations remains a priority.

### *Model Dependence*

The patient simulator and AI Decision Aid both use GPT-4.1, though information leakage between them is structurally prevented. The simulator's internal profile and concept markers are stripped before its response reaches the AI Decision Aid, which extracts clinical information from natural language text alone, as it would from a real patient. Shared model family could still produce correlated blind spots not present with two independently trained systems. The LLM judge (Claude Opus 4.6) is a different model, mitigating this risk at the evaluation layer but not for the simulator-decision aid pairing. Future work should test decision aid performance against simulator output using an alternative model family on a subsample.

### *Annotation Diversity*

All annotations were performed by two raters; while their complementary clinical backgrounds (psychology, nursing) and adjudicated agreement provide some assurance, a two-rater design limits the generalizability of the reported agreement estimates and cannot capture the full range of expert interpretation. Future evaluations should incorporate a larger, more diverse annotator pool.

## Conclusions

Applied to a conversational antidepressant decision aid, our patient simulation framework found a steep gap in performance across health literacy levels. Rank-1 concept retrieval was 81.9% for proficient health literacy and 47.6% for limited literacy. Recommendation accuracy followed the same pattern. This gap matters, as limited health literacy is common among patients with the most severe psychiatric illness. The patients least well served by this system are those who most need it to work. Identifying this risk required simulated patients grounded across all three dimensions: medical, linguistic, and behavioral. Prior simulators have varied one or two of these, but not all three together. Though we evaluated the framework only on antidepressant selection, extending it to other clinical decision-aid tasks is future work. It contributes a risk-ratio-based algorithm for auditable medical profile generation, a controlled perturbation method for validating annotators, and an LLM judge with human-level agreement. It also enables scalable simulation and evaluation, so candidate mitigations can be developed and tested against the same risks before deployment. But technical fixes for AI systems are not enough. Health systems must specify care pathway, oversight, and escalation requirements; they must also validate the system clinically in the populations it underserves. Code, source data, and conversation logs are publicly available.

## Acknowledgments

The study used data from the All of Us Research Program's Registered Tier Dataset v8, available to all authorized users on the Researcher Workbench. We gratefully acknowledge All of Us participants for their contributions, without whom this research would not have been possible. We also thank the National Institutes of Health's All of Us Research Program for making available the participant data examined in this study.

We also thank the members of the project advisory board, comprising clinicians, leaders of national mental health organizations, and individuals with lived experience of depression, for their ongoing guidance throughout the PCORI-funded research program.

We thank Francisca Mainoo, MPH, BSN, RN and Roya Minovi, BA for their careful annotation work supporting the evaluation reported in this study.

The authors used Claude Opus 4.7 (Anthropic) and GPT-5.5 (OpenAI), accessed through their respective web interfaces, for language polishing during manuscript preparation. The tools were not used to generate scientific content, perform analysis, or produce references. The authors reviewed all suggestions and retain full responsibility for the manuscript.

## Funding

This work was supported by a Patient-Centered Outcomes Research Institute (PCORI) Award [ME-2024C1-36732]. The views in this article are solely the responsibility of the authors and do not necessarily represent the views of the PCORI, its Board of Governors or Methodology Committee.

## Conflicts of Interest

FA has a pending patent (#19/253,342) assigned to George Mason University related to the DEMI methods described in this work. The other authors declare no conflicts of interest.

## Data Availability

Medical profiles were generated using risk-ratio gating based on EHR data from the All of Us Research Program Registered Tier Dataset v8. Antidepressant response probabilities were computed using the DEMI algorithm based on the same dataset. The risk-ratio database and DEMI model weights will be released with publication (participant-level All of Us records cannot be redistributed under the program's data use agreement). All code, simulated patient profiles, conversation logs, and evaluation annotations are available on GitHub [24]

## Authors' Contributions

**MTRS:** Conceptualization, Data curation, Formal analysis, Methodology, Investigation, Resources, Validation, Visualization, Writing – original draft. **MSI:** Data curation, Formal analysis, Resources, Validation. **HRAE:** Data curation, Writing – review & editing. **KRR:** Data curation, Formal analysis, Validation. **YL:** Data curation, Formal analysis, Validation. **VFC:** Data curation, Project administration, Resources, Validation, Writing – review & editing. **KPE:** Conceptualization, Funding acquisition, Investigation, Writing – review & editing. **FA:** Conceptualization, Data curation, Formal analysis, Methodology, Funding acquisition, Investigation, Project administration, Supervision, Validation, Writing – review & editing. **KL:** Conceptualization, Data curation, Formal analysis, Methodology, Funding acquisition, Investigation, Project administration, Supervision, Validation, Writing – original draft.

All authors reviewed and approved the final manuscript.

## Abbreviations

AI RMF: Artificial Intelligence Risk Management Framework

CoT: Chain-of-Thought

CPT-4: Current Procedural Terminology, 4th edition

DEMI: Direct Effects Multiplicative Inference

DSM: Diagnostic and Statistical Manual of Mental Disorders

EHR: Electronic Health Record

FKGL: Flesch-Kincaid Grade Level

IAA: Inter-Annotator Agreement

JSON: JavaScript Object Notation

LLM: Large Language Model

NIST: National Institute of Standards and Technology

PCORI: Patient-Centered Outcomes Research Institute

RAG: Retrieval-Augmented Generation

RR: Risk Ratio

SNOMED CT: Systematized Nomenclature of Medicine Clinical Terms

t-SNE: t-distributed stochastic neighbor embedding

**Multimedia Appendix 1:** Patient Simulator

**Multimedia Appendix 2:** Medical and Behavioral Profiles

**Multimedia Appendix 3:** Retrieval Performance

## References

1. Laranjo L, Dunn AG, Tong HL, et al. Conversational agents in healthcare: a systematic review. *J Am Med Inform Assoc*. 2018;25(9):1248-1258. doi:10.1093/jamia/ocy072. PMID:30010941

2. Maity S, Saikia MJ. Large language models in healthcare and medical applications: A review. *Bioengineering*. 2025;12(6):631. doi:10.3390/bioengineering12060631. PMID:40564447

3. Thirunavukarasu AJ, Ting DSJ, Elangovan K, Gutierrez L, Tan TF, Ting DSW. Large language models in medicine. *Nat Med*. 2023;29(8):1930-1940. doi:10.1038/s41591-023-02448-8

4. Tu T, Schaekermann M, Palepu A, et al. Towards conversational diagnostic artificial intelligence. *Nature*. 2025;642(8067):442-450. doi:10.1038/s41586-025-08866-7. PMID:40205050

5. Kwame A, Petrucka PM. A literature-based study of patient-centered care and communication in nurse-patient interactions: barriers, facilitators, and the way forward. *BMC Nurs*. 2021;20:158. doi:10.1186/s12912-021-00684-2. PMID:34479560

6. Wynia MK, Osborn CY. Health literacy and communication quality in health care organizations. *J Health Commun*. 2010;15(Suppl 2):102-115. doi:10.1080/10810730.2010.499981. PMID:20845197

7. Gourabathina A, Gerych W, Pan E, Ghassemi M. The medium is the message: How non-clinical information shapes clinical decisions in LLMs. In: *Proceedings of the 2025 ACM Conference on Fairness, Accountability, and Transparency*. ACM; 2025:1805-1828. doi:10.1145/3715275.3732121

8. Hua Y, Xia W, Bates D, et al. Standardizing and scaffolding health care AI-chatbot evaluation: Systematic review. *JMIR AI*. 2025;4(1):e69006. doi:10.2196/69006

9. Templin T, Fort S, Padmanabham P, et al. Framework for bias evaluation in large language models in healthcare settings. *Npj Digit Med*. 2025;8(1):414. doi:10.1038/s41746-025-01786-w

10. Nadarzynski T, Knights N, Husbands D, et al. Achieving health equity through conversational AI: A roadmap for design and implementation of inclusive chatbots in healthcare. *PLOS Digit Health*. 2024;3(5):e0000492. doi:10.1371/journal.pdig.0000492

11. Wornow M, Xu Y, Thapa R, et al. The shaky foundations of large language models and foundation models for electronic health records. *Npj Digit Med*. 2023;6:135. doi:10.1038/s41746-023-00879-8. PMID:37516790

12. Guan S, Xiong H, Wang J, Bian J, Zhu B, Lou JG. Evaluating LLM-based agents for multi-turn conversations: A survey. *arXiv*. Preprint posted online March 28, 2025:arXiv:2503.22458. doi:10.48550/arXiv.2503.22458

13. Kwan WC, Zeng X, Jiang Y, et al. MT-Eval: A multi-turn capabilities evaluation benchmark for large language models. In: *Proceedings of the 2024 Conference on Empirical Methods in Natural Language Processing*. Association for

Computational Linguistics; 2024:20153-20177. doi:10.18653/v1/2024.emnlp-main.1124

14. Balog K, Zhai C. User simulation for evaluating information access systems. In: *Proceedings of the Annual International ACM SIGIR Conference on Research and Development in Information Retrieval in the Asia Pacific Region*. SIGIR-AP '23. Association for Computing Machinery; 2023:302-305. doi:10.1145/3624918.3629549

15. Kalinich M, Luccarelli J, Moss F, Torous J. Leveraging simulation to provide a practical framework for assessing the novel scope of risk of LLMs in healthcare. *medRxiv*. Preprint posted online November 13, 2025:2025.11.10.25339903. doi:10.1101/2025.11.10.25339903

16. Tabassi E. Artificial intelligence risk management framework (AI RMF 1.0). *NIST*. Published online January 26, 2023. Accessed October 28, 2025. https://www.nist.gov/publications/artificial-intelligence-risk-management-framework-ai-rmf-10

17. All of Us Research Program Investigators, Denny JC, Rutter JL, et al. The "All of Us" research program. *N Engl J Med*. 2019;381(7):668-676. doi:10.1056/NEJMsr1809937. PMID:31412182

18. Institute of Medicine (US) Committee on Health Literacy. *Health Literacy: A Prescription to End Confusion*. (Nielsen-Bohlman L, Panzer AM, Kindig DA, eds.). National Academies Press (US); 2004. Accessed February 8, 2026. http://www.ncbi.nlm.nih.gov/books/NBK216032/. PMID:25009856

19. Nutbeam D. Health literacy as a public health goal: a challenge for contemporary health education and communication strategies into the 21st century. *Health Promot Int*. 2000;15(3):259-267. doi:10.1093/heapro/15.3.259

20. Pennebaker JW, Mehl MR, Niederhoffer KG. Psychological aspects of natural language use: Our words, our selves. *Annu Rev Psychol*. 2003;54:547-577. doi:10.1146/annurev.psych.54.101601.145041

21. Rude S, Gortner EM, Pennebaker J. Language use of depressed and depression-vulnerable college students. *Cogn Emot*. 2004;18(8):1121-1133. doi:10.1080/02699930441000030

22. Roter D, Hall JA. *Doctors Talking with Patients/Patients Talking with Doctors*. Bloomsbury Publishing; 2006.

23. Street RL, Makoul G, Arora NK, Epstein RM. How does communication heal? Pathways linking clinician–patient communication to health outcomes. *Patient Educ Couns*. 2009;74(3):295-301. doi:10.1016/j.pec.2008.11.015

24. lybargerlanguagelab/Patient-Simulation. Published online 2026. Accessed August 19, 2026. https://github.com/lybargerlanguagelab/Patient-Simulation

25. Gao C, Lan X, Li N, et al. Large language models empowered agent-based modeling and simulation: a survey and perspectives. *Humanit Soc Sci Commun*. 2024;11(1):1259. doi:10.1057/s41599-024-03611-3

26. Montaner M, López B, de la Rosa JL. Evaluation of recommender systems through simulated users. In: *6th International Conference on Enterprise Information Systems*. 2004:303-308. Accessed March 18, 2026. https://www.scitepress.org/Link.aspx?doi=10.5220/0002622703030308

27. Yao S, Halpern Y, Thain N, et al. Measuring recommender system effects with simulated users. *arXiv*. Preprint posted online January 12, 2021:arXiv:2101.04526. doi:10.48550/arXiv.2101.04526

28. Luo X, Tang Z, Wang J, Zhang X. DuetSim: Building user simulator with dual large language models for task-oriented dialogues. In: Calzolari N, Kan MY, Hoste V, Lenci A, Sakti S, Xue N, eds. *Proceedings of the 2024 Joint International Conference on Computational Linguistics, Language Resources and Evaluation (LREC-COLING 2024)*. ELRA and ICCL; 2024:5414-5424. Accessed August 14, 2025. https://aclanthology.org/2024.lrec-main.481/

29. Schatzmann J, Weilhammer K, Stuttle M, Young S. A survey of statistical user simulation techniques for reinforcement-learning of dialogue management strategies. *Knowl Eng Rev*. 2006;21(2):97-126. doi:10.1017/S0269888906000944

30. Labhishetty S. *Models and Evaluation of User Simulation in Information Retrieval*. Thesis. University of Illinois at Urbana-Champaign; 2023. Accessed November 29, 2025. https://hdl.handle.net/2142/120103

31. Zhang Y, Liu X, Zhai C. Information retrieval evaluation as search simulation: A general formal framework for IR evaluation. In: *Proceedings of the ACM SIGIR International Conference on Theory of Information Retrieval*. ICTIR ’17. Association for Computing Machinery; 2017:193-200. doi:10.1145/3121050.3121070

32. Yun T, Yang E, Safdari M, et al. Sleepless nights, sugary days: Creating synthetic users with health conditions for realistic coaching agent interactions. In: Che W, Nabende J, Shutova E, Pilehvar MT, eds. *Findings of the Association for Computational Linguistics: ACL 2025*. Association for Computational Linguistics; 2025:14159-14181. doi:10.18653/v1/2025.findings-acl.729

33. Cho YM, Rai S, Ungar L, Sedoc J, Guntuku SC. An integrative survey on mental health conversational agents to bridge computer science and medical

perspectives. *Proc Conf Empir Methods Nat Lang*. 2023;2023:11346-11369. doi:10.18653/v1/2023.emnlp-main.698. PMID:38618627

34. Cook DA, Overgaard J, Pankratz VS, Del Fiol G, Aakre CA. Virtual patients using large language models: Scalable, contextualized simulation of clinician-patient dialogue with feedback. *J Med Internet Res*. 2025;27:e68486. doi:10.2196/68486. PMID:39854611

35. Kyung D, Chung H, Bae S, et al. PatientSim: A persona-driven simulator for realistic doctor-patient interactions. *arXiv*. Preprint posted online October 29, 2025:arXiv:2505.17818. doi:10.48550/arXiv.2505.17818

36. Liu L, Yang X, Lei J, et al. A survey on medical large language models: Technology, application, trustworthiness, and future directions. *arXiv*. Preprint posted online December 9, 2024:arXiv:2406.03712. doi:10.48550/arXiv.2406.03712

37. Holderried F, Stegemann-Philipps C, Herrmann-Werner A, et al. A language model–powered simulated patient with automated feedback for history taking: Prospective study. *JMIR Med Educ*. 2024;10:e59213. doi:10.2196/59213. PMID:39150749

38. Luo MJ, Bi S, Pang J, et al. A large language model digital patient system enhances ophthalmology history taking skills. *Npj Digit Med*. 2025;8(1):502. doi:10.1038/s41746-025-01841-6

39. Walonoski J, Kramer M, Nichols J, et al. Synthea: An approach, method, and software mechanism for generating synthetic patients and the synthetic electronic health care record. *J Am Med Inform Assoc*. 2018;25(3):230-238. doi:10.1093/jamia/ocx079

40. Dahmen J, Cook D. SynSys: A synthetic data generation system for healthcare applications. *Sensors*. 2019;19(5):1181. doi:10.3390/s19051181

41. Yan C, Zhang Z, Nyemba S, Li Z. Generating synthetic electronic health record data using generative adversarial networks: Tutorial. *JMIR AI*. 2024;3:e52615. doi:10.2196/52615. PMID:38875595

42. Yoon J, Mizrahi M, Ghalaty NF, et al. EHR-Safe: generating high-fidelity and privacy-preserving synthetic electronic health records. *Npj Digit Med*. 2023;6(1):141. doi:10.1038/s41746-023-00888-7

43. Chen X, Wu Z, Shi X, Cho H, Mukherjee B. Generating synthetic electronic health record data: a methodological scoping review with benchmarking on phenotype data and open-source software. *J Am Med Inform Assoc*. 2025;32(7):1227-1240. doi:10.1093/jamia/ocaf082

44. Rabaey P, Heytens S, Demeester T. SimSUM: Simulated benchmark with structured and unstructured medical records. *arXiv*. Preprint posted online July 10, 2025:arXiv:2409.08936. doi:10.48550/arXiv.2409.08936

45. Yu H, Zhou J, Li L, et al. Simulated patient systems are intelligent when powered by large language model-based AI agents. *arXiv*. Preprint posted online July 29, 2025:arXiv:2409.18924. doi:10.48550/arXiv.2409.18924

46. Bodonhelyi A, Stegemann-Philipps C, Sonanini A, et al. Modeling challenging patient interactions: LLMs for medical communication training. *arXiv*. Preprint posted online March 2025:arXiv:2503.22250. doi:10.48550/arXiv.2503.22250

47. Wang R, Milani S, Chiu JC, et al. PATIENT-ψ: Using large language models to simulate patients for training mental health professionals. In: Al-Onaizan Y, Bansal M, Chen YN, eds. *Proceedings of the 2024 Conference on Empirical Methods in Natural Language Processing*. Association for Computational Linguistics; 2024:12772-12797. doi:10.18653/v1/2024.emnlp-main.711

48. Bao Z, Liu Q, Huang X, Wei Z. SFMSS: Service flow aware medical scenario simulation for conversational data generation. In: Chiruzzo L, Ritter A, Wang L, eds. *Findings of the Association for Computational Linguistics: NAACL 2025*. Association for Computational Linguistics; 2025:4586-4604. doi:10.18653/v1/2025.findings-naacl.259

49. Sehgal NKR, Kambhamettu H, Chang A, Zhu A, Ungar L, Guntuku SC. PAL: Designing conversational agents as scalable, cooperative patient simulators for palliative-care training. In: *Companion Publication of the 2025 Conference on Computer-Supported Cooperative Work and Social Computing*. ACM; 2025:346-350. doi:10.1145/3715070.3749250

50. Wojtusiak J. *Towards Intelligent Patient Data Generator*. Machine Learning and Inference Laboratory, George Mason University; 2016. Accessed March 18, 2026. https://www.mli.gmu.edu/papers/2016/16-9.pdf

51. Hua Y, Siddals S, Ma Z, et al. Charting the evolution of artificial intelligence mental health chatbots from rule-based systems to large language models: a systematic review. *World Psychiatry*. 2025;24(3):383-394. doi:10.1002/wps.21352

52. Alemi F, Aljuaid M, Durbha N, et al. A surrogate measure for patient reported symptom remission in administrative data. *BMC Psychiatry*. 2021;21:121. doi:10.1186/s12888-021-03133-1. PMID:33663440

53. Top 10 Most Common Antidepressants Dispensed in the U.S. Definitive Healthcare. Accessed November 11, 2025. https://www.definitivehc.com/resources/healthcare-insights/top-antidepressants-by-prescription-volume

54. Lewis AE, Weiskopf N, Abrams ZB, et al. Electronic health record data quality assessment and tools: a systematic review. *J Am Med Inform Assoc*. 2023;30(10):1730-1740. doi:10.1093/jamia/ocad120

55. Si Y, Du J, Li Z, et al. Deep representation learning of patient data from Electronic Health Records (EHR): A systematic review. *J Biomed Inform*. 2021;115:103671. doi:10.1016/j.jbi.2020.103671

56. Syed R, Eden R, Makasi T, et al. Digital health data quality issues: Systematic review. *J Med Internet Res*. 2023;25:e42615. doi:10.2196/42615. PMID:37000497

57. Lim E, He YV, Joselowitz J, et al. MATRIX: Multi-Agent simulaTion fRamework for safe Interactions and conteXtual clinical conversational evaluation. *arXiv*. Preprint posted online August 26, 2025:arXiv:2508.19163. doi:10.48550/arXiv.2508.19163

58. Moëll B, Aronsson FS. Harm reduction strategies for thoughtful use of large language models in the medical domain: Perspectives for patients and clinicians. *J Med Internet Res*. 2025;27(1):e75849. doi:10.2196/75849

59. Sørensen K, Van den Broucke S, Fullam J, et al. Health literacy and public health: a systematic review and integration of definitions and models. *BMC Public Health*. 2012;12:80. doi:10.1186/1471-2458-12-80. PMID:22276600

60. Zhou X, Kim H, Brahman F, et al. HAICOSYSTEM: An ecosystem for sandboxing safety risks in human-AI interactions. *arXiv*. Preprint posted online August 30, 2025:arXiv:2409.16427. doi:10.48550/arXiv.2409.16427

61. Guyon I, Elisseeff A. An introduction to variable and feature selection. *J Mach Learn Res*. 2003;3(2003):1157-1182.

62. Peng H, Long F, Ding C. Feature selection based on mutual information criteria of max-dependency, max-relevance, and min-redundancy. *IEEE Trans Pattern Anal Mach Intell*. 2005;27(8):1226-1238. doi:10.1109/TPAMI.2005.159

63. Cevasco KE, Morrison Brown RE, Woldeselassie R, Kaplan S. Patient engagement with conversational agents in health applications 2016–2022: A systematic review and meta-analysis. *J Med Syst*. 2024;48(1):40. doi:10.1007/s10916-024-02059-x. PMID:38594411

64. Wolf MS, Bailey SC. The role of health literacy in patient safety. *PSNet Internet*. Published online March 22, 2009. Accessed February 11, 2026. https://psnet.ahrq.gov/perspective/role-health-literacy-patient-safety

65. Lee S, Kim M, Cherif L, et al. Learning diverse attacks on large language models for robust red-teaming and safety tuning. *arXiv*. Preprint posted online May 28, 2024:arXiv:2405.18540. doi:10.48550/arXiv.2405.18540

66. Perez E, Huang S, Song F, et al. Red teaming language models with language models. In: *Proceedings of the 2022 Conference on Empirical Methods in Natural Language Processing*. Association for Computational Linguistics; 2022:3419-3448. doi:10.18653/v1/2022.emnlp-main.225

67. CDC. National Action Plan to Improve Health Literacy. Health Literacy. October 7, 2024. Accessed November 4, 2025. https://www.cdc.gov/health-literacy/php/develop-plan/national-action-plan.html

68. Paasche-Orlow MK, Wolf MS. The causal pathways linking health literacy to health outcomes. *Am J Health Behav*. 2007;31 Suppl 1:S19-26. doi:10.5555/ajhb.2007.31.supp.S19. PMID:17931132

69. Tausczik YR, Pennebaker JW. The psychological meaning of words: LIWC and computerized text analysis methods. *J Lang Soc Psychol*. 2010;29(1):24-54. doi:10.1177/0261927X09351676

70. DeVault D, Artstein R, Benn G, et al. SimSensei kiosk: a virtual human interviewer for healthcare decision support. In: *Proceedings of the 2014 International Conference on Autonomous Agents and Multi-Agent Systems*. AAMAS '14. International Foundation for Autonomous Agents and Multiagent Systems; 2014:1061-1068. Accessed March 18, 2026. https://dl.acm.org/doi/10.5555/2615731.2617415

71. Fitzpatrick KK, Darcy A, Vierhile M. Delivering cognitive behavior therapy to young adults with symptoms of depression and anxiety using a fully automated conversational agent (Woebot): A randomized controlled trial. *JMIR Ment Health*. 2017;4(2):e7785. doi:10.2196/mental.7785

72. Liao Y, Meng Y, Wang Y, Liu H, Wang Y, Wang Y. Automatic interactive evaluation for large language models with state aware patient simulator. *arXiv*. Preprint posted online July 21, 2024:arXiv:2403.08495. doi:10.48550/arXiv.2403.08495

73. Sanjeewa R, Iyer R, Apputhurai P, Wickramasinghe N, Meyer D. Empathic conversational agent platform designs and their evaluation in the context of mental health: Systematic review. *JMIR Ment Health*. 2024;11(1):e58974. doi:10.2196/58974

74. Simpson S, McDowell A. *The Clinical Interview: Skills for More Effective Patient Encounters*. Routledge; 2019. doi:10.4324/9780429437243

75. Wei J, Wang X, Schuurmans D, et al. Chain-of-thought prompting elicits reasoning in large language models. In: *Proceedings of the 36th International Conference on Neural Information Processing Systems*. NIPS '22. Curran Associates Inc.; 2022:24824-24837.

76. Alemi F, Lin Y, Elyazori HRA, Cardenas VF, Ramezani N, Lybarger K. Causal AI reasoning: Direct effects multiplicative inference algorithm. *SSRN*. Preprint posted online 2026. doi:10.2139/ssrn.6537875

77. Kincaid JP, Fishburne J, Rogers RL, Chissom BS. Derivation of new readability formulas (automated readability index, fog count and Flesch reading ease formula) for Navy enlisted personnel. Published online February 1, 1975:RBR875. Accessed February 10, 2026. https://apps.dtic.mil/sti/html/tr/ADA006655/

78. Loper E, Bird S. NLTK: The natural language toolkit. In: *Proceedings of the ACL-02 Workshop on Effective Tools and Methodologies for Teaching Natural Language Processing and Computational Linguistics*. Association for Computational Linguistics; 2002:63-70. doi:10.3115/1118108.1118117

79. Salazar A. XLM-RoBERTa Depression Detection Model. Published online 2025. https://github.com/malexandersalazar/xlm-roberta-base-cls-depression

80. Bevendorff J, Casals XB, Chulvi B, et al. Overview of PAN 2024: Multi-author writing style analysis, multilingual text detoxification, oppositional thinking analysis, and generative AI authorship verification — Extended abstract. In: Goharian N, Tonellotto N, He Y, et al., eds. *Advances in Information Retrieval - 46th European Conference on Information Retrieval, ECIR 2024, Glasgow, UK, March 24-28, 2024, Proceedings, Part VI*. Vol 14613. Lecture Notes in Computer Science. Springer; 2024:3-10. doi:10.1007/978-3-031-56072-9_1

81. Bhat S, Varma V. Large language models as annotators: A preliminary evaluation for annotating low-resource language content. In: Deutsch D, Dror R, Eger S, et al., eds. *Proceedings of the 4th Workshop on Evaluation and Comparison of NLP Systems*. Association for Computational Linguistics; 2023:100-107. doi:10.18653/v1/2023.eval4nlp-1.8

82. Pavlovic M, Poesio M. The effectiveness of LLMs as annotators: A comparative overview and empirical analysis of direct representation. In: Abercrombie G, Basile V, Bernadi D, et al., eds. *Proceedings of the 3rd Workshop on Perspectivist Approaches to NLP (NLPerspectives) @ LREC-COLING 2024*. ELRA and ICCL; 2024:100-110. Accessed February 11, 2026. https://aclanthology.org/2024.nlperspectives-1.11/

83. Zhang R, Li Y, Ma Y, Zhou M, Zou L. LLMaAA: Making large language models as active annotators. *arXiv*. Preprint posted online October 31, 2023:arXiv:2310.19596. doi:10.48550/arXiv.2310.19596

84. Du X, Zhou Z, Wang Y, et al. Testing and evaluation of generative large language models in electronic health record applications: a systematic review. *J Am Med Inform Assoc*. 2026;33(3):743-753. doi:10.1093/jamia/ocaf233

85. Abbo ED, Zhang Q, Zelder M, Huang ES. The increasing number of clinical items addressed during the time of adult primary care visits. *J Gen Intern Med*. 2008;23(12):2058-2065. doi:10.1007/s11606-008-0805-8. PMID:18830762

86. Chase H. LangChain. Published online 2022. Accessed May 7, 2026. https://github.com/langchain-ai/langchain

87. Jurafsky D, Martin JH. *Speech and Language Processing: An Introduction to Natural Language Processing, Computational Linguistics, and Speech Recognition with Language Models*. 3rd ed. 2026. https://web.stanford.edu/~jurafsky/slp3/

88. Trivedi MH, Rush AJ, Wisniewski SR, et al. Evaluation of outcomes with citalopram for depression using measurement-based care in STAR*D: implications for clinical practice. *Am J Psychiatry*. 2006;163(1):28-40. doi:10.1176/appi.ajp.163.1.28. PMID:16390886

89. Rush AJ. Challenges of research on treatment-resistant depression: a clinician's perspective. *World Psychiatry*. 2023;22(3):415-417. doi:10.1002/wps.21136

90. Degan TJ, Kelly PJ, Robinson LD, Deane FP, Smith AM. Health literacy of people living with mental illness or substance use disorders: A systematic review. *Early Interv Psychiatry*. 2021;15(6):1454-1469. doi:10.1111/eip.13090

91. Paasche-Orlow MK, Parker RM, Gazmararian JA, Nielsen-Bohlman LT, Rudd RR. The prevalence of limited health literacy. *J Gen Intern Med*. 2005;20(2):175-184. doi:10.1111/j.1525-1497.2005.40245.x

92. Lorant V. Socioeconomic inequalities in depression: A meta-analysis. *Am J Epidemiol*. 2003;157(2):98-112. doi:10.1093/aje/kwf182

93. Linder A, Gerdtham UG, Trygg N, Fritzell S, Saha S. Inequalities in the economic consequences of depression and anxiety in Europe: a systematic scoping review. *Eur J Public Health*. 2020;30(4):767-777. doi:10.1093/eurpub/ckz127

94. Bennabi D, Vandel P, Papaxanthis C, Pozzo T, Haffen E. Psychomotor retardation in depression: A systematic review of diagnostic, pathophysiologic, and therapeutic implications. *BioMed Res Int*. 2013;2013:1-18. doi:10.1155/2013/158746

95. Pan Z, Park C, Brietzke E, et al. Cognitive impairment in major depressive disorder. *CNS Spectr*. 2019;24(1):22-29. doi:10.1017/S1092852918001207

96. Raja S, Hasnain M, Hoersch M, Gove-Yin S, Rajagopalan C. Trauma informed care in medicine: Current knowledge and future research directions. *Fam Community Health*. 2015;38(3):216-226. doi:10.1097/FCH.0000000000000071

97. Mihelicova M, Brown M, Shuman V. Trauma-informed care for individuals with serious mental illness: An avenue for community psychology's involvement in community mental health. *Am J Community Psychol*. 2018;61(1-2):141-152. doi:10.1002/ajcp.12217

98. Rush AJ, Trivedi MH, Wisniewski SR, et al. Acute and longer-term outcomes in depressed outpatients requiring one or several treatment steps: A STAR*D report. *Am J Psychiatry*. 2006;163(11):1905-1917. doi:10.1176/ajp.2006.163.11.1905

99. Obermeyer Z, Powers B, Vogeli C, Mullainathan S. Dissecting racial bias in an algorithm used to manage the health of populations. *Science*. 2019;366(6464):447-453. doi:10.1126/science.aax2342

100. Srivarathan A, Bradford A, Shearkhani S, et al. Bridging diagnostic safety and mental health: a systematic review highlighting inequities in autism spectrum disorder diagnosis. *BMJ Qual Saf*. Published online August 25, 2025:bmjqs-2025-018723. doi:10.1136/bmjqs-2025-018723

101. Lincoln A, Espejo D, Johnson P, et al. Limited literacy and psychiatric disorders among users of an urban safety-net hospital's mental health outpatient clinic. *J Nerv Ment Dis*. 2008;196(9):687-693. doi:10.1097/NMD.0b013e31817d0181

102. Chowdhary N, Jotheeswaran AT, Nadkarni A, et al. The methods and outcomes of cultural adaptations of psychological treatments for depressive disorders: a systematic review. *Psychol Med*. 2014;44(6):1131-1146. doi:10.1017/S0033291713001785

103. Chong WW, Aslani P, Chen TF. Effectiveness of interventions to improve antidepressant medication adherence: a systematic review: Effectiveness of interventions to improve antidepressant adherence. *Int J Clin Pract*. 2011;65(9):954-975. doi:10.1111/j.1742-1241.2011.02746.x

104. Osterberg L, Blaschke T. Adherence to medication. *N Engl J Med*. 2005;353(5):487-497. doi:10.1056/NEJMra050100

105. Kroenke K, Spitzer RL, Williams JBW. The PHQ-9: Validity of a brief depression severity measure. *J Gen Intern Med*. 2001;16(9):606-613. doi:10.1046/j.1525-1497.2001.016009606.x

106. Hamilton M. A rating scale for depression. *J Neurol Neurosurg Psychiatry*. 1960;23(1):56-62. doi:10.1136/jnnp.23.1.56

# Multimedia Appendix 1: Patient Simulator

## NIST AI RMF Alignment

The patient simulator design is grounded in the NIST AI Risk Management Framework (AI RMF) [1], which defines four core functions: *GOVERN*, *MAP*, *MEASURE*, and *MANAGE*. This work aligns with the MAP and MEASURE functions through three profile dimensions, each addressing a distinct category of risk. Medical profiles target clinical accuracy risks arising from high-dimensional EHR data, linguistic profiles target communication-dependent risks emerging from health literacy variation, and behavioral profiles target interaction-dependent risks from adversarial and disengaged patient behavior. Table 7 maps the five AI RMF-aligned trustworthiness requirements to their corresponding data challenges and mitigated risks.

Table 7: Medical profile generation requirements: NIST alignment, challenges addressed, and risks mitigated.

| **Requirement (NIST Functions)** | **Data Challenge** | **AI Risk** |
|---|---|---|
| Controllability (MAP, MEASURE) | High-dimensional feature space with sparse task-relevant signal | Spurious associations and overfitting; inflated performance estimates |
| Coherence (MAP, MEASURE) | Coding inconsistencies and implausible feature combinations | Medically contradictory profiles masking safety vulnerabilities |
| Variability (MAP, MEASURE) | Skewed distributions and underrepresentation | Subgroup disparities and incomplete risk detection |
| Efficiency (GOVERN, MEASURE) | Excessive record density limiting evaluation scale | Evaluation bottlenecks and reduced oversight capacity |
| Transparency (GOVERN, MEASURE) | Obscured feature provenance and latent interactions | Limited auditability and hidden bias propagation |

## Architecture and System Prompt

The Patient Simulator architecture operationalizes the generation of psychologically realistic patient responses through the coordinated integration of medical, linguistic, and behavioral profiles. This modular design ensures that each simulated interaction remains clinically grounded, linguistically differentiated, and behaviorally consistent. Figure 7 shows the structure of the system prompt used to enforce this alignment, with a brief description of its components provided below.

Concretely, the system prompt is organized around three profile components that constrain content, expression, and interaction behavior:

1. **Medical Profile:** The simulator uses a structured medical history represented by hierarchical indices (e.g., [2.3], [3.1]), where each code corresponds to a specific clinical fact or condition. These indices are crucial for traceability and annotation, allowing evaluators or downstream models to link simulated responses back to precise clinical data points.
2. **Linguistic Profile:** The linguistic profile defines how medical facts are style-transferred to reflect the patient's expressive characteristics such as tone, vocabulary, and sentence complexity. This ensures that the same underlying medical information can appear differently across patients (e.g., formal vs. colloquial language) while maintaining factual alignment.
3. **Behavioral Profile:** The behavioral layer determines how the simulated patient engages with the conversation, controlling elements such as adherence, emotional engagement, topical focus, and adversarial tendencies. By enforcing behavioral constraints, the simulator produces psychologically credible dialogue patterns ranging from cooperative to avoidant or argumentative behaviors.

   **Question-Response Logic:** For each intake question, the simulator follows a structured reasoning pipeline:

   a. Identify Relevant Medical Facts: Locate facts from the medical profile that relate to the current query, referencing them by their index (e.g., [3.2] Depression diagnosis).

   b. Apply Style Transfer: For each fact, rephrase the medical term or phrase according to the linguistic profile while preserving its factual meaning.

   c. Construct a Natural Response: Integrate the style-transferred facts into a contextually appropriate, natural-language answer that aligns with both linguistic and behavioral traits. Each style-transferred phrase is wrapped in <\s>...<\s> for traceability and followed by its [X.Y] reference tag.

**Response Format:**

All outputs adhere to a strict JSON schema to ensure transparency, reproducibility, and structured analysis. This output format is essential for the annotation process, as it allows annotators to review each conversational turn, examine the corresponding medical history, and verify the accuracy and consistency of the patient simulator's responses. The output JSON contains:

- "relevant_medical_history" – the original indexed facts used in the response.
- "style_transferred_medical_history" – the linguistically modified versions of those facts.

"response" – the final natural-language utterance embedding both versions for clarity and interpretability.

You are simulating a psychologically realistic patient based on three profiles: Medical, Behavioral, and Linguistic.

1. **Medical Profile:**

    Demographics:

        1.1: Age: 34

        1.2: Gender: Male

        ... ... ...

    Diagnosis History:

        2.1: Generalized Anxiety Disorder

        2.2: Hyperlipidemia variants

        ... ... ...

    Procedures:

        3.1: Psychiatric Diagnostic Interview Examination

        3.2: Individual Psychotherapy

        ... ... ...

2. **Linguistic Profile:**

    **Style:** Concrete, informal, sometimes vague.

    **Tone:** Hesitant, uncertain, conversational.

    **Vocab:** Everyday terms, slang, vague quantities.

    **Structure:** Short, fragmented sentences; frequent fillers.

    **Patterns:** Minimal elaboration unless prompted.

3. **Behavioral Profile:**

    **Conversational Adherence:** Follows prompts; stays structured.

    **Engagement:** Highly engaged; adds helpful details.

    **Topical Focus:** Stays on topic; avoids tangents.

    **Adversarial/Toxic Behavior:** None; consistently polite.

**For each of the questions:**

➔ **Step 1:** Identify relevant medical facts using their assigned index (e.g., [3.2] Individual Psychotherapy).

➔ **Step 2:** For each identified fact, apply style transfer according to the linguistic profile.

➔ **Step 3:** Construct a natural language response that embeds the style-transferred facts embedding the linguistic and behavioral profile. Each style-transferred phrase must be **wrapped in `<\s>` ... `<\s>`** and must also include the **[X.Y]** reference right after it.

**Response format:**

    "relevant_medical_history": [

        "[1.2] Gender: Male",

        "[3.2] Individual Psychotherapy"]

    "style_transferred_medical_history": [

        "[1.2] male",

        "[3.2] talked to someone"],

    "response": "Final response using the above style-transferred facts with inline references and <\s> tags.

Figure 7: Prompt Structure of Patient Simulator

## References

1. Tabassi E. Artificial intelligence risk management framework (AI RMF 1.0). *NIST*. Published online January 26, 2023. Accessed October 28, 2025. https://www.nist.gov/publications/artificial-intelligence-risk-management-framework-ai-rmf-10

# Multimedia Appendix 2: Medical and Behavioral Profiles

## Medical Profiles

**Phase – 1: Medical Profile Generation**

This section presents the medical profile generation procedure used by the analytical advice system in detail. The method follows a staged selection strategy that balances outcome relevance, statistical coherence, and controlled diversity to produce clinically plausible and informative profiles.

Outcome Relevance (Top-K Filter): The algorithm first restricts the candidate space to the Top-K predictors of the antidepressant response outcome $e_0$ (here, K = 500). This filter improves sample efficiency and signal-to-noise by focusing subsequent selection on variables with demonstrated individual association to $e_0$. As in classical filter style feature selection, an initial relevance screen improves downstream generalization and reduces variance in later steps. In spirit, this aligns with the "max-relevance" component of minimum-redundancy-maximum-relevance (mRMR), which explicitly prioritizes features that are informative about the target before addressing redundancy.

Coherence and Approximate Independence via a Symmetric Risk-Ratio Gate: New candidates are screened using a symmetric acceptance band around unity based on aggregated pairwise risk ratios with respect to the already-selected set S. Let $RR(s, v)$ denote the risk ratio between an existing event $s \in S$ and a candidate event $v$, and let $aggregate_{s \in S}$ be a conservative aggregator (e.g., MAX) over these pairwise values. Acceptance condition:

$$1/1.5 < aggregate_{s \in S}\, RR(s, v) \leq high$$

This symmetric band ($low = 1/1.5$) respects the multiplicative interpretation of $RR$ treating protective associations ($< 1$) and harmful associations ($> 1$) in a balanced manner while capping near deterministic couplings ($> high$) that would introduce redundancy or multicollinearity. Excluding values $\ll 1$ also avoids strongly anticorrelated additions that can create unstable compensations in downstream predictions [1]. Using MAX as the aggregator is conservative (rejects $v$ if it couples too strongly with any $s \in S$). Controlled Diversity from the Residual Event Pool: After establishing the coherent core of relevant events (Top-K features), the algorithm introduces controlled diversity by exploring the residual event pool. Those features not included among the top-ranked relevant set. In this phase, a residual candidate $u \in R$ is admitted only if it exhibits a meaningful positive association with at least one of the already selected relevant features $s \in S$. This step aims to capture latent or contextually related variables that were not strongly ranked individually but still contribute to plausible patient variability when correlated with the existing core.

To ensure statistical coherence, independence is defined as the range $1/1.5 \leq RR(s,v) \leq 1.5$; any relationship within this interval is treated as neutral and thus ignored. Accordingly, only candidates showing stronger associations i.e., $RR(s,v) > 1.5$ or $RR(s,v) < 1/1.5$ are eligible for inclusion, provided they were not part of the original relevant set. Diversity condition:

$$\exists s \in S : RR(s,v) > 1.5$$

where R denotes the residual pool (features outside the Top-500), and S represents the previously selected relevant features. This mechanism ensures that the resulting simulated patient profiles include a bounded degree of heterogeneity driven by correlated but previously under-ranked features, enriching diversity without compromising coherence [1,2]

The Risk Ratio categories and the corresponding feature inclusion logic are illustrated in Figure 8, which depicts the transition from negatively associated to positively associated features and their mapping across Step 3 (relevant) and Step 4 (coherent but irrelevant) feature sets.

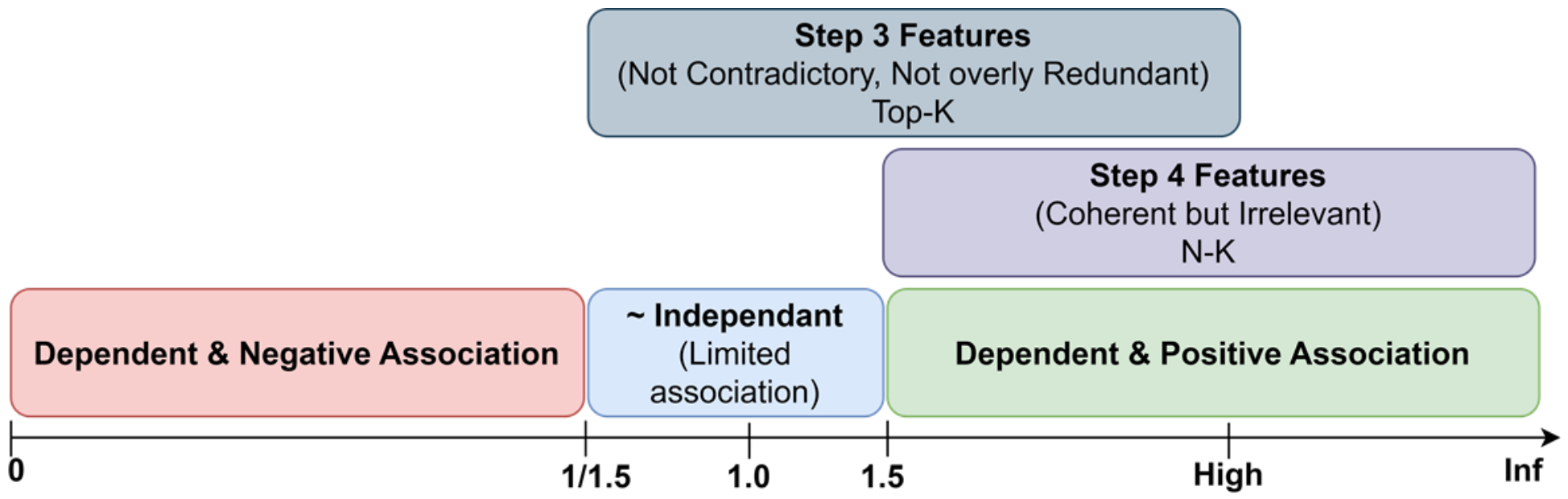

Figure 8: Graphical representation of Risk Ratio thresholds and feature selection workflow.

The medical profile generation algorithm constructs a simulated clinical history by selecting events that are both clinically meaningful and statistically consistent with real world response patterns. The process begins with a set of variables that strongly predict response to the target antidepressant, then filters these candidates through risk based constraints to ensure independence and realistic variability. A second expansion stage introduces additional events that provide controlled diversity without creating improbable clinical combinations. The final set of events forms a structured medical profile that reflects both individual level variability and population level statistical associations.

The inputs listed below define the core elements required for the generation of each simulated medical profile. They specify the clinical target, the size of the simulated population, the available concept codes, and the demographic distributions from which each patient begins, along with the parameters that control risk ratio constraints and diversity expansion.

Inputs:

- e0 – Concept code of response of target antidepressant
- N– Number of patients to simulate
- All_Concept_Code – List of all the concept code
- G – Categorical distribution over gender (At Birth) codes
- A – Categorical distribution over age-bin codes
- high – Maximum allowed Risk Ratio for selecting a variable
- max_residual_additions – Limit on how many residual-pool events may be appended

The helper routines provide the key decision checks used throughout the selection process. They evaluate the compatibility of candidate events with the partially constructed profile and ensure that all additions remain within the required risk ratio bounds. These routines allow the algorithm to enforce statistical constraints without interrupting the main flow of the pseudocode.

**Helper Routines:**

- $PassesRRGate(selected, candidate_event, low, high)$ - Returns $TRUE$ if the aggregated pairwise risk ratio $RR(s, candidate_event)$ across all s in selected satisfies $1/1.5 < aggregate \leq high$. Otherwise returns $FALSE$.
- $AnyRRExceeds(selected, candidate_event, threshold)$ - Returns $TRUE$ if there exists at least one event s in selected such that $RR(s, candidate_event)$ exceeds the given threshold. Otherwise returns $FALSE$.
- $PredictResp(S, e_0)$ : Returns the probability of response to the antidepressant associated with concept $e_0$, given the selected event set $S$, computed using the DEMI algorithm.

Figure 9 combines the inputs and helper routines into a four-stage algorithm that builds each medical profile step by step. It begins with a Top-K predictor filter and initializes demographic seeds, then applies an independence screen governed by high, the upper similarity threshold used in the RR gate, to ensure statistical coherence, and finally introduces controlled diversity through residual events capped by max_residual_additions, which can be randomly drawn from a range (e.g., 3-5) rather than fixed. Together, these stages create a clinically plausible and statistically grounded event set for each simulated patient, with high and max_residual_additions offering two tunable levers for controlling profile density:

lowering high narrows the set of eligible candidates, while max_residual_additions directly sets how many additional features are added.

**Procedure:**

```
# Step-1: Selection of Top-K and Initialization
1. V ← Top500_Predictor_of_e_0(e_0).        \\ Top_500 predictors
2. R ← {All_Concept_Code} \ {V ∪ G ∪ A}    \\ Residual Event
3. Initialize output container T ← ∅.
4. For each patient j = 1..N:
5.     S ← ∅   // selected set.
6.     A_v ← V;  A_r ← R   // fresh available pools.
# Step-2: Demographics Seed
7.     S ← S ∪ {weighted_random_choice(G)}   \\ gender
8.     S ← S ∪ {weighted_random_choice(A)}   \\ age-bin
# Step-3: Independence-Screened Selection
9.     While A_v ≠ ∅:
10.          v ← random_choice(A_v).
11.          If PassesRRGate(S, v, high=high, low=1/1.5):
12.                Set S ← S ∪ {v}.
13.          Remove v from A_v.
14.    Define S_intermediate ← S.
# Step-4: Controlled Diversity Expansion
15.    i ← 1.
16.    While A_r ≠ ∅ and i ≤ max_residual_additions:
17,           u ← random_choice(A_r).
18.          If AnyRRExceeds(S_intermediate, u, threshold = 1.5):
19.                Set S ← S ∪ {u};  i ← i + 1.
20.          Remove u from A_r.
21.    Compute final p^ ← PredictResp(S, e_0).
22.    Append (j, e_0, |S|, S, p^) to T.
```

Figure 9. Medical profile generation algorithm

## Phase – 2: Probabilistic Patient Selection Using Binomial Distribution

The second phase of the medical profile generation process selects a subset of simulated patients whose response probabilities collectively reproduce the population-level response distribution observed in the reference dataset. Each simulated patient from Phase 1 is associated with a predicted probability of antidepressant response between 0 and 1. The selection aims to retain a cohort whose aggregate distribution mirrors the expected binomial pattern defined by the population mean response rate ($p$) and total number of patients ($n$).

To operationalize this, the probability range [0, 1] is divided into seven probability bands corresponding to approximate σ-intervals of a binomial distribution $B(n, p)$. Each simulated patient is assigned to a band according to their predicted probability of response. The population mean ($\mu = np$) and standard deviation ($\sigma = \sqrt{(np(1-p))}$ ) define the expected frequency of patients in each band, which acts as a target allocation weight. For example, for $n = 100$ and $p = 0.4$, the binomial distribution has $\mu = 40$ and $\sigma \approx 4.9$; approximately 64 % of the population lies within $(\mu - \sigma, \mu + \sigma]$, 15 % on each side within $(\mu \pm 1\sigma\ to\ \mu \pm 2\sigma]$, and only a few percent beyond $\mu \pm 2\sigma$. Table 8 shows a generalized sigma-band allocation for the binomial distribution $B(n, p)$, including the corresponding ranges and their interpretations.

Table 8: Generalized Sigma-Band Allocation for Binomial Distribution $B\ (n, p)$.

| **Sigma Band Interval** | **Range of Success Counts (k)** | **Approximate Probability Mass (%)** | **Interpretation** |
|---|---|---|---|
| k ≤ μ − 3σ | 0 ≤ k ≤ ⌊μ − 3σ⌋ | ≈ 0.1–0.3 % | Extremely low outcome region |
| (μ − 3σ, μ − 2σ] | ⌊μ − 3σ⌋ + 1 ≤ k ≤ ⌊μ − 2σ⌋ | ≈ 2–3 % | Moderately below average region |
| (μ − 2σ, μ − 1σ] | ⌊μ − 2σ⌋ + 1 ≤ k ≤ ⌊μ − 1σ⌋ | ≈ 14–16 % | Slightly below average region |
| (μ − 1σ, μ + 1σ] | ⌊μ − 1σ⌋ + 1 ≤ k ≤ ⌊μ + 1σ⌋ | ≈ 63–65 % | Central or typical outcome region |
| (μ + 1σ, μ + 2σ] | ⌊μ + 1σ⌋ + 1 ≤ k ≤ ⌊μ + 2σ⌋ | ≈ 14–16 % | Slightly above average region |
| (μ + 2σ, μ + 3σ] | ⌊μ + 2σ⌋ + 1 ≤ k ≤ ⌊μ + 3σ⌋ | ≈ 2–3 % | Moderately high outcome region |
| k > μ + 3σ | ⌊μ + 3σ⌋ + 1 ≤ k ≤ n | ≈ 0.1–0.3 % | Extremely high outcome region |

Patients are then sampled proportionally to these weights so that the empirical distribution of predicted response probabilities in the final cohort approximates the theoretical binomial distribution. This approach ensures that the selected subset preserves both the central tendency and natural variability of the population while avoiding over- or under-representation of extreme responders. The outcome is a statistically balanced and demographically diverse cohort suitable for downstream simulation and evaluation.

## Behavioral Profiles

We present the mapping of 13 commonly observed patient behaviors documented by Simpson et al. [3] into four higher-level behavioral categories*: Inquisitive and Open-Ended, Reserved & Minimalist*, *Distracted & Unfocused*, and *Adversarial & Combative* in Table 9. In addition to these four categories, *Structured & Cooperative* is included as the baseline behavioral category.

Table 9: Mapping of behavioral profiles to their associated patient behaviors.

| **Behavioral Profile** | **Associated Patient Situation** |
|---|---|
| Inquisitive & Open Ended | Is uncertain what they want |
| Reserved & Minimalist | Is reluctant to work together<br>Is anxious, worried, or upset<br>Worries they will never get better |
| Adversarial & Combative | Is angry<br>Is demanding<br>Brings up negative feelings in the clinician |
| Distracted & Unfocused | Is difficult to direct or disorganized<br>Has an unclear diagnosis<br>Is an unreliable historian<br>Has a difficult time changing their behavior<br>Does not take medications regularly<br>Has difficulty planning ahead |

## References

1. Peng H, Long F, Ding C. Feature selection based on mutual information criteria of max-dependency, max-relevance, and min-redundancy. *IEEE Trans Pattern Anal Machine Intell*. 2005;27(8):1226-1238. doi:10.1109/TPAMI.2005.159

2. Guyon I, Elisseeff A. An introduction to variable and feature selection. *Journal of machine learning research*. 2003;3(2003):1157-1182.

3. Simpson S, McDowell A. *The Clinical Interview: Skills for More Effective Patient Encounters*. Routledge; 2019. doi:10.4324/9780429437243

## Multimedia Appendix 3: Retrieval Performance

Table 10 details retrieval performance by linguistic profile over all reference concepts.

Table 10: *AI Decision Aid* Retrieval Performance for Identified Concepts (Reported over All Reference Concepts)

| **Linguistic Profile** | | **n** | **Rank-1** | **Top-20** | **Not Retrieved** | **Mean Rank (non-Rank-1)** |
|---|---|---|---|---|---|---|
| **Health Literacy** | Limited | 460 | 47.6% | 71.5% | 12.0% | 14.5 |
| | Functional | 1023 | 69.6% | 83.1% | 8.8% | 8.5 |
| | Proficient | 453 | 81.9% | 85.2% | 10.2% | 9.2 |
| **Condition-specific** | Depression | 440 | 62.0% | 79.5% | 9.8% | 9.6 |
| | Illness Anxiety Disorder | 443 | 63.7% | 77.9% | 9.9% | 13.0 |
| | **Overall** | **2,819** | **65.9%** | **80.2%** | **9.9%** | **11.2** |

Figure 10 breaks down Rank-1 and Rank-20 accuracy across linguistic profiles, behavioral profiles, and their intersections. The health literacy gradient is visible across all behavioral conditions, with *Limited* health literacy consistently producing the lowest Rank-1 accuracy and the widest gap between Rank-1 and Rank-20, indicating that correct concepts are retrieved but fail to rank first.

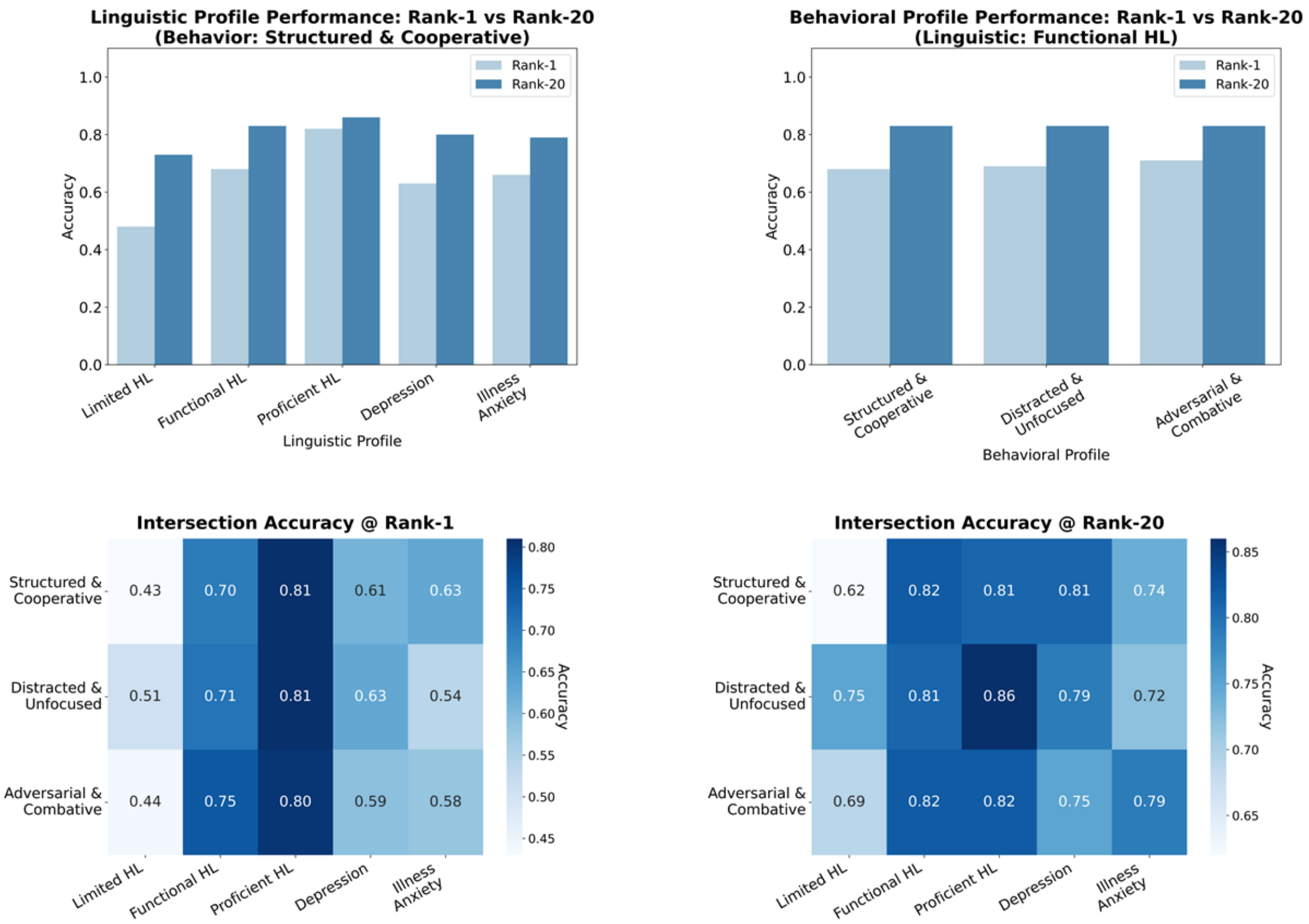

Figure 10: Rank-1 and Rank-20 accuracy for linguistic profiles, behavioral profiles, and their intersections.